\pdfoutput=1
\documentclass{article}

    \PassOptionsToPackage{numbers, compress}{natbib}

\usepackage[final]{neurips_data_2024}

\usepackage[utf8]{inputenc} 
\usepackage[T1]{fontenc}    
\usepackage{hyperref}       
\usepackage{url}            
\usepackage{booktabs}       
\usepackage{amsfonts}       
\usepackage{nicefrac}       
\usepackage{microtype}      
\usepackage{xcolor}         
\usepackage{pifont}

\usepackage{subcaption}
\usepackage{wrapfig}
\usepackage{multirow}
\usepackage{makecell}
\usepackage{xspace}
\usepackage{enumitem}
\usepackage{float}
\usepackage{amsmath}
\usepackage{listings}
\usepackage{algorithm}
\usepackage{algorithmic}
\usepackage{amsfonts,amssymb}
\usepackage{mathrsfs}
\usepackage{subcaption}
\usepackage{graphicx}       
\usepackage{longtable}
\usepackage{threeparttable}
\usepackage{siunitx}
\usepackage[normalem]{ulem}
\useunder{\uline}{\ul}{}

\usepackage{minitoc}
\newcommand{\hhide}[1]{}
\newcommand{\model}[0]{\textsc{VisualAgentBench}\xspace}
\newcommand{\sm}[0]{\textsc{VAB}\xspace}
\newcommand{\vpara}[1]{\noindent\textbf{#1}\xspace} %

\newcommand{\problem}[0]{LMM-as-Visual-Foundation-Agent\xspace}

\newcommand{\greencheck}{{\color{green}\ding{51}}} 
\newcommand{\redx}{{\color{red}\ding{55}}} 
\newcommand\ystar{\raisebox{-2pt}{\includegraphics[width=0.8em]{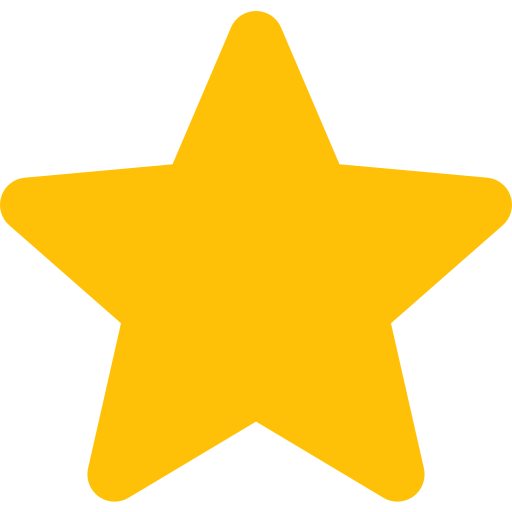}}}

\definecolor{deepgreen}{RGB}{0,120,0}
\definecolor{deepred}{RGB}{180,0,0}

\newcommand{\pscore}[2]{%
  \begin{minipage}[t]{2.3cm}
    \centering #1  
    \raggedleft \textcolor{deepgreen}{\scriptsize $\uparrow$#2\%} 
  \end{minipage}%
}

\newcommand{\dscore}[2]{%
  \begin{minipage}[t]{2.3cm}
    \centering #1  
    \raggedleft \textcolor{deepred}{\scriptsize $\downarrow$#2\%} 
  \end{minipage}%
}

\usepackage{authblk}

\makeatletter
\renewcommand\AB@affilsepx{, \protect\Affilfont}
\makeatother

\title{VisualAgentBench: Towards Large Multimodal Models as Visual Foundation Agents}

\author[1,*]{{Xiao Liu}}
\author[3,$\dagger$,*]{{Tianjie Zhang}}
\author[2,*]{{Yu Gu}}
\author[1]{{Iat Long Iong}}
\author[1]{{Yifan Xu}}
\author[1]{{Xixuan Song}}
\author[1]{{Shudan Zhang}}
\author[1]{{Hanyu Lai}}
\author[1,$\dagger$]{{Xinyi Liu}}
\author[1,$\dagger$]{{Hanlin Zhao}}
\author[1,$\dagger$]{{Jiadai Sun}}
\author[1,$\dagger$]{{Xinyue Yang}}
\author[1,$\dagger$]{{Yu Yang}}
\author[1]{{Zehan Qi}}
\author[1,$\dagger$]{{Shuntian Yao}}
\author[1]{{Xueqiao Sun}}
\author[4,$\dagger$]{{Siyi Cheng}}
\author[1]{{Qinkai Zheng}}
\author[1]{{Hao Yu}}
\author[1]{{Hanchen Zhang}}
\author[1]{{Wenyi Hong}}
\author[1]{{Ming Ding}}
\author[1]{{Lihang Pan}}
\author[1]{{Xiaotao Gu}}
\author[1]{{Aohan Zeng}}
\author[1]{{Zhengxiao Du}}
\author[2]{Chan Hee Song}
\author[2]{Yu Su}
\author[1]{{Yuxiao Dong}}
\author[1]{{Jie Tang}}
\affil[1]{Tsinghua University}
\affil[2]{The Ohio State University}
\affil[3]{Zhejiang University}
\affil[4]{Peking University}

\begin{document}

\doparttoc
\faketableofcontents

\maketitle
\renewcommand{\thefootnote}{\fnsymbol{footnote}}
    \footnotetext[1]{Equal contribution. Email: \texttt{\{shawliu9,mistyreed63849\}@gmail.com, gu.826@osu.edu}}
    \footnotetext[2]{Work done when these authors visited Tsinghua University.}
\renewcommand{\thefootnote}{\arabic{footnote}}
\vspace{-8.5mm}
\begin{abstract}
\vspace{-2mm}
  Large Multimodal Models (LMMs) have ushered in a new era in artificial intelligence, merging capabilities in both language and vision to form highly capable \textbf{Visual Foundation Agents}.
  These agents are postulated to excel across a myriad of tasks, potentially approaching general artificial intelligence. 
  However, existing benchmarks fail to sufficiently challenge or showcase the full potential of LMMs in complex, real-world environments. 
  To address this gap, we introduce VisualAgentBench (\sm), a comprehensive and pioneering benchmark specifically designed to train and evaluate LMMs as visual foundation agents across diverse scenarios, including Embodied, Graphical User Interface, and Visual Design, with tasks formulated to probe the depth of LMMs’ understanding and interaction capabilities. 
  Through rigorous testing across nine proprietary LMM APIs and eight open models, we demonstrate the considerable yet still developing agent capabilities of these models. 
  Additionally, \sm constructs a trajectory training set constructed through hybrid methods including Program-based Solvers, LMM Agent Bootstrapping, and Human Demonstrations, promoting substantial performance improvements in LMMs through behavior cloning. 
  Our work not only aims to benchmark existing models but also provides a solid foundation for future development into visual foundation agents.
  Code, train \& test data, and part of fine-tuned open LMMs are available at \url{https://github.com/THUDM/VisualAgentBench}.
\end{abstract}

\begin{figure}[h]
    \centering
    \vspace{-8mm}
    \begin{subfigure}{0.42\textwidth}
        \centering
        \includegraphics[width=\linewidth]{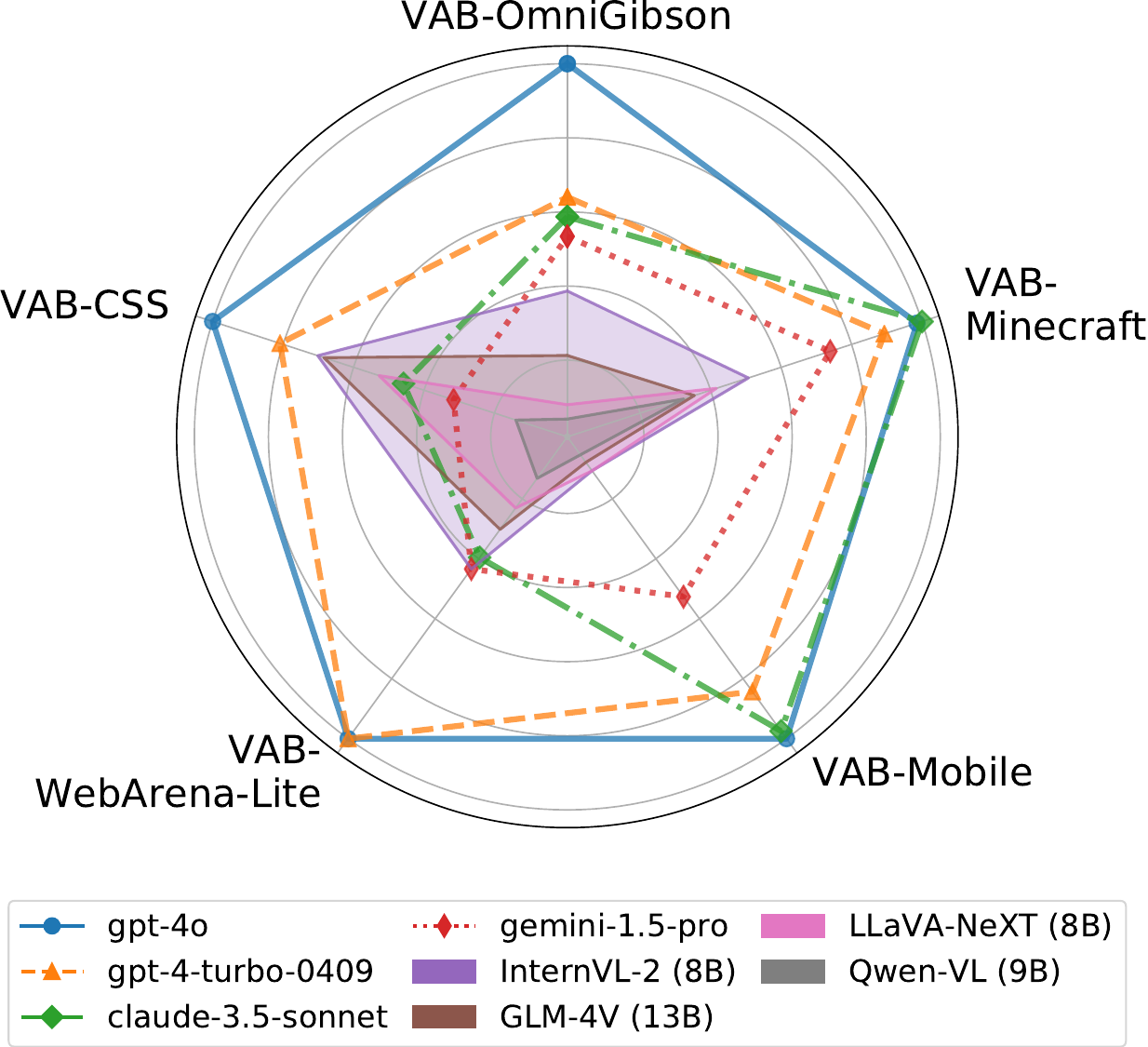}
        \caption{Typical LMMs’ \sm performance (relative) against the best in each environment.}
    \end{subfigure}
    \hspace{.5mm}
    \begin{subfigure}{0.51\textwidth}
        \centering
        \includegraphics[width=\linewidth]{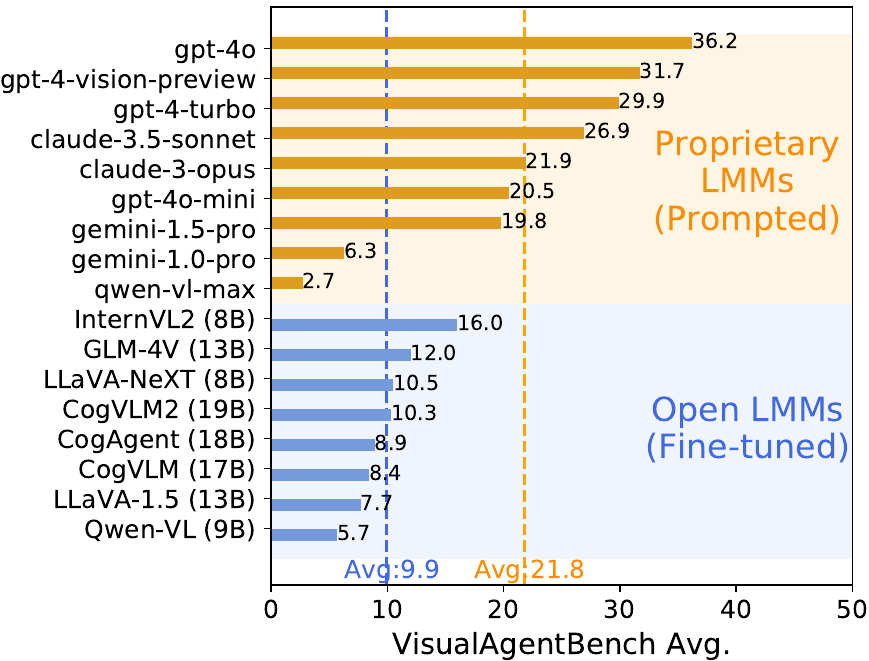}
        \caption{Average \sm Success Rates of tested LMMs across 5 environments. Dashed lines for two LMM types’ average.}
    \end{subfigure}
    \vspace{-2mm}
    \caption{Overview of Proprietary and Open LMMs on \model. After Behavior Cloning (BC) on trajectories, Open LMMs demonstrate potential to serve as visual foundation agents.}
    \label{fig:intro}
    \vspace{-6mm}
\end{figure}

\begin{figure}[t]
    \centering
    \includegraphics[width=\textwidth]{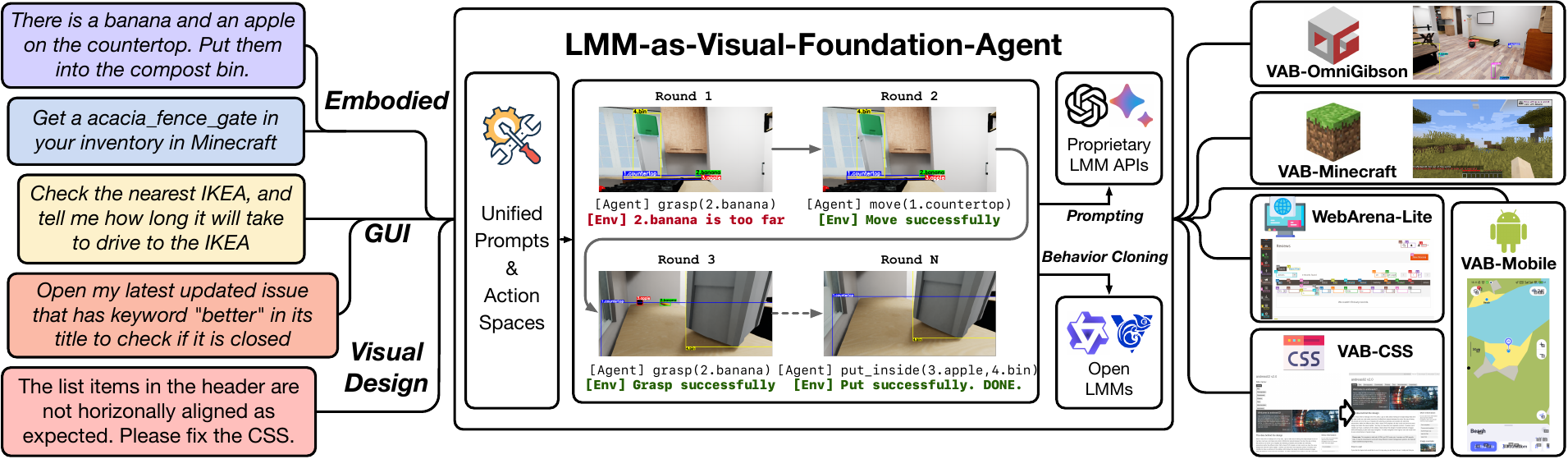}
    \caption{\model is the first systematic benchmark to evaluate LMM-as-Visual-Agent across a diverse set of practical challenges. Beside benchmarking, it is also the first to offer SFT trajectory data for behavior cloning training over all target environments, demonstrating the huge potential to improve open LMMs for serving as visual foundation agents.}
    \label{fig:main}
    \vspace{-6mm}
\end{figure}

\section{Introduction}

Recent advancements in Foundation Models, particularly Large Language Models (LLMs) \cite{GPT3,chowdhery2022palm,touvron2023llama,zeng2022glm} and Large Multimodal Models (LMMs) \cite{liu2024visual,GPT-4-vision-preview,GPT-4o,Claude-3}, have showcased their profound capabilities in understanding and processing vast amounts of world knowledge, factual information, and common sense reasoning. 
Notably, these models have demonstrated potential as intelligent agents \cite{Searle1970SpeechAA,Maes1994AgentsTR,wooldridge1995intelligent}, addressing a broad spectrum of real-world challenges \cite{liu2023agentbench}. 
LMMs, in particular, enhance the capabilities of these agents by integrating visual inputs, thereby expanding the scope of intelligent agent applications.

This progress has given rise to the concept of \textbf{Foundation Agents}—generalist agents adept at mastering a plethora of skills across various virtual and embodied environments, mirroring human versatility. 
These agents, especially those powered by LMMs, are envisioned to excel in multitask environments without the need for task-specific fine-tuning, a paradigm already set by LLM-based language agents. 
The burgeoning field of visual foundation agents offers promising pathways toward achieving AGI, with the potential to significantly elevate human productivity and creativity.

However, the setup for \textbf{\problem} remains underdeveloped. 
Most existing evaluations on LMMs focus on traditional tasks like Visual Question Answering (VQA) \cite{kembhavi2016diagram,singh2019towards,lu2022learn}, Optical Character Recognition (OCR) \cite{liu2023hidden}, and Referring Expression Generation (REG) \cite{kazemzadeh2014referitgame}, or on performance in standardized human exams \cite{yue2023mmmu,lu2023mathvista}. 
These assessments rarely measure the models' higher-level reasoning and planning capabilities or their specific strengths as visual agents. 
In contrast, the role of LLMs as agents in text environments has been extensively explored and validated as a reliable measure of their capabilities \cite{yao2022react,liu2023agentbench}.

Recent benchmarks for multimodal agents, while valuable, do not adequately address the comprehensive evaluation required for \problem. 
These benchmarks often limit their focus to single environments such as Household \cite{shridhar2020alfred,shridhar2020alfworld}, Gaming \cite{fan2022minedojo,wu2023smartplay}, Web \cite{deng2024mind2web,zhou2023webarena,koh2024visualwebarena}, or Desktop scenarios \cite{xie2024osworld,kapoor2024omniact}. 
This narrow scope prevents a holistic assessment of LMMs’ multitask agent capabilities. 
Furthermore, the prevalent prompting-only evaluation in existing benchmarks does not suffice for open LMMs \cite{liu2024visual,bai2023qwen,wang2023cogvlm}, which typically show limited instruction-following capabilities so far, thus hindering a comprehensive evaluation.

To bridge this gap, we introduce \model (\sm)---the first systematic benchmark designed to multitask train and evaluate visual foundation agents across a diverse array of realistic vision-centric tasks. 
We present three representative scenarios and develop five distinct datasets for this study: \textbf{Embodied} (\sm-OmniGibson, \sm-Minecraft), \textbf{Graphical User Interface (GUI)} (\sm-Mobile~\cite{anonymous2024android}, \sm-WebArena-lite \cite{zhou2023webarena}), and \textbf{Visual Design} (\sm-CSS), enabling comprehensive testing and development of agents that can navigate complex spaces, interact with digital interfaces, and understand aesthetic and functional aspects of visual design. 
This diversity not only challenges the agents’ capabilities across different settings but also enhances their adaptability and utility in practical applications, paving the way for more robust and versatile visual foundation agents.

We have standardized the prompting and data formats to facilitate a consistent evaluation of visual foundation agents across these environments. 
Each \sm task is assessed through interactive evaluation \cite{liu2023agentbench,zhou2023webarena,jimenez2023swe,yao2022webshop}, where LMMs engage directly with the environment, and their performance is measured by specific judge functions. 
This approach presents practical challenges for LMM agents, rendering \sm a more robust and realistic benchmark compared to those relying on offline trajectories \cite{deng2024mind2web,rawles2024androidinthewild}.
Our extensive testing, with nine proprietary LMM APIs and eight open LMMs using \sm, demonstrates the impressive progress of \problem.
Top proprietary LMMs, such as \texttt{gpt-4o}, are solving \num{36.2}\% of challenging problems with mere prompting.
Nonetheless, it is also shown that their performances are still far from being practically deployable.

Significantly, \sm also includes a training set comprising \num{4482} ground truth trajectories across five environments, curated through a blend of \textbf{Program-based Solvers}, \textbf{LMM Agent Bootstrapping}, and \textbf{Human Demonstrations}. 
We explore methodologies for employing and integrating these strategies based on environmental characteristics. 
Our experiments demonstrate that behavior cloning (BC) on the \sm training set markedly enhances the capabilities of open LMMs as visual agents, with most surpassing the performance of proprietary LMMs like \texttt{gemini-1.0-pro} and \texttt{qwen-vl-max}, and approaching close towards \texttt{gemini-1.5-pro}.
Nevertheless, the gaps between top proprietary LMMs and open models could be wide, as top proprietary LMMs can still significantly outperform finetuned open models with mere prompting.

In summary, our contributions are threefold:
\begin{itemize}[leftmargin=1.5em,itemsep=0pt,parsep=0.2em,topsep=0.0em,partopsep=0.0em]
\item The introduction of \sm, a pioneering benchmark for both training and evaluating visual agents across diverse and realistic challenges, featuring five datasets and three key scenarios. We have standardized prompting and data formatting to streamline the assessment of foundation agents across various environments.
\item The development of a hybrid data curation pipeline to construct the \sm training set, containing \num{4482} high-quality training trajectories from five environments. Our findings indicate that behavior cloning on these trajectories substantially improves the performance of open LMMs.
\item A comprehensive evaluation of nine proprietary LMM APIs and eight open LMMs using \sm, revealing significant insights into the current state and potential of LMMs as visual agents across multiple domains. Our results highlight the substantial performance gaps between proprietary and open models and suggest directions for future research in visual foundation agents.
\end{itemize}

\section{Problem Formulation and \sm Design Features}
\label{section:design}
In this section, we introduce the problem definition of \problem.
Upon the definition, we explain a series of practical principles we follow during the design of \sm.

\vpara{\problem.}
An agentic problem could be generally formulated as a Partially Observable Markov Decision Process (POMDP) problem, where $\mathcal{S}$ denotes the state space, $\mathcal{A}$ denotes the action space, $\mathcal{T}$ denotes the transition function, $\mathcal{R}$ refers to the reward function, $\mathcal{I}$ refers to the instruction space, and $\mathcal{O}$ refers to the observation space.
Compared to LLM-as-Agent~\cite{liu2023agentbench}, the observation space $\mathcal{O}$ must incorporate visual inputs (e.g., images or videos) in \problem, significantly extending the application scope but also casting a substantial challenge for LMMs to reconcile their multimodal understanding and high-level reasoning.

\vpara{Design Features of \sm.}
Given that LMMs are still evolving rapidly, we adhere to several principles in our design of \sm to accommodate the current capabilities and limitations of LMMs.
\begin{itemize}[leftmargin=1.5em,itemsep=0pt,parsep=0.2em,topsep=0.0em,partopsep=0.0em]
    \item \textbf{Vision-Centric:} \sm agent tasks are designed to primarily rely on visual inputs to solve problems. While additional text inputs could be beneficial, \sm aims to evaluate how LMMs perform when perceiving the environment as humans do in agent tasks. For example, while HTML is shown useful for Web GUI Agent~\cite{zhou2023webarena,deng2024mind2web}, humans typically browse the internet from screens without reading HTMLs.
    \item \textbf{High-Level Decision Making:} \sm focuses on evaluating LMMs' high-level decision-making abilities. Compared to prior smaller visual-language models that specifically target low-level policies~\cite{lynch2020language,brohan2022rt,lifshitz2024steve}, LMMs excel at high-level planning and interacting~\cite{driess2023palm} in text response thanks to their commonsense, knowledge, and flexible instruction following with mere prompting. Therefore, in \sm, we simplify the low-level control by providing convenient action interfaces, and ask tested LMMs to concentrate on delivering high-level decision sequences in text. 
    \item \textbf{Interactive Evaluation:} Evaluating LLMs or LMMs on real-world agent tasks is challenging, as task goals can be achieved by various means. As a result, it becomes a mainstream practice to evaluate in an interactive manner~\cite{liu2023agentbench,zhou2023webarena,jimenez2023swe,xie2024osworld}. \sm also adheres to this principle.
    \item \textbf{Trajectories for Behavior Cloning:} Many previous execution-based agent benchmarks for LLMs and LMMs, despite being realistic and challenging, often fail to provide effective training sets for the community to use for improvement. LLMs and LMMs need behavior cloning training on trajectories for better performance~\cite{nakano2021webgpt,zeng2023agenttuning,lai2024autowebglm}. However, creating such datasets consisting of valid instructions, trajectories, and reward functions is costly and requires annotators' good understanding of the environment. In response to the challenge, for each \sm environment we endeavor to deliver instructions created with a hybrid set of strategies (Cf. Section~\ref{subsection:training}). Experiments show that our constructed training sets can effectively improve the performance of open LMMs on \sm.
\end{itemize}
Note that as the field advances, some of the above principles may become obsolete and irrelevant.
We will continuously update \sm to accommodate the progress of LMMs.
\section{\model: Tasks and Environments}

In \sm, we carefully select the most representative and promising tasks that could be enabled by LMM-based agents.
These tasks generally fall into three categories: embodied agents, including household and game environments; GUI agents, covering mobile and web apps; and visual design agents, focusing on frontend CSS debugging (Figure~\ref{fig:main}). 
They span diverse domains and feature unique challenges, providing an ideal testbed for a comprehensive evaluation of LMM-based agents.
When constructing \sm, we strictly follow the principles outlined in Section~\ref{section:design}.
Our efforts focus on addressing gaps in evaluating LMM-based agents while leveraging existing resources to avoid redundancy, ensuring all our work is meaningful and avoids reinventing the wheel.
For \num{4} out of \num{5} tasks, we collect new data from scratch.
For web agents, we adapt and clean WebArena~\cite{zhou2023webarena} as our test set, as it is already suitable for LMM-based evaluation.
For household agents, we use the OmniGibson environment from Behavior-1k \cite{li2023behavior} and create new tasks based on high-level actions we defined, which are crucial for evaluating LMM-based agents and absent in existing datasets.
We similarly construct our tasks in Minecraft using the MineRL environment\footnote{\url{https://minerl.readthedocs.io}} with our self-defined high-level actions.
Finally, for our mobile app and CSS debugging tasks, we create new interactive environments due to the lack of suitable resources in the literature and collect data based on these environments.
An overview of \sm is shown in Table~\ref{table:related}.

\subsection{Embodied Agent}
Embodied agents have been a central topic in AI, naturally involving multimodal sensory data, including language and vision signals. 
The multimodal capabilities of LMMs could enable new possibilities for embodied agents.

\vpara{\sm-OmniGibson.}
One of the most actively researched environments in embodied AI is the household environment due to its complexity and range of everyday tasks~\cite{huang2022language,llm-planner,shridhar2020alfred}.
We build the household environment for embodied agents using OmniGibson, a high-fidelity simulator based on Nvidia Omniverse that features diverse scenes and realistic physical effects.\footnote{\url{https://www.nvidia.com/en-us/omniverse/}}
An example activity in \sm-OmniGibson would be ``\textit{Put all 8 plates from the countertops into the cabinet in the kitchen}'', where agents should accomplish the tasks using provided high-level actions (e.g.,``\texttt{grasp}'', ``\texttt{put\_inside}''). 
We adopt the task \textit{success rate} as the evaluation metric. (Cf. Appendix~\ref{appendix:omnigibson}).

\vpara{\sm-Minecraft.}
Minecraft has become a popular open-world environment for developing generalist embodied agents due to its diverse tasks (e.g., survival, harvesting, crafting, combat, and creative tasks), varied environments, and interactive mobs, necessitating generalized agent abilities~\cite{fan2022minedojo,lifshitz2024steve}. 
In \sm-Minecraft, the agent is expected to accomplish a wide range of tasks, including item collection and killing hostile mobs.
An example task in \sm-Minecraft would be ``\textit{Get a fishing rod in your inventory}'', and the LMM agent need to interact with the game environment using provided scripts (e.g.,``\texttt{craft}'', ``\texttt{smelt}'') or calling a low-level controller Steve-1~\cite{lifshitz2024steve} with prompt. We adopt the task \textit{success rate} as the evaluation metric. (Cf. Appendix~\ref{appendix:minecraft})

\subsection{GUI Agent}
GUI is another typical scenario where LMM agents may excel.
Compared to embodied environments, GUI environments are more information-intensive and require a good understanding of UI elements and layouts.
We provide two interactive and reproducible GUI environments, Mobile (i.e., Android) and WebArena, to evaluate LMM GUI agents in a practical manner.

\vpara{\sm-Mobile~\cite{anonymous2024android}.}
Automated agents on Android GUI can significantly advance personal digital assistants. Although pioneer works like MOTIF~\cite{burns2022dataset} and AITW~\cite{rawles2024androidinthewild} have explored training and evaluating these agents, they typically use Step Success Rate evaluated offline. Recent works~\cite{yang2023appagent,wang2024mobile} leverage LMMs as Android GUI agents but lack reproducible executive evaluation frameworks. We address this by creating tasks for LMM agents to perform human-like actions (e.g., \texttt{Tap}, \texttt{Swipe}) on smartphones using Android Virtual Device (AVD). For example, ``\textit{Find a hotpot restaurant nearby and make a reservation for me tonight}.'' Agents must understand the Android GUI and make decisions based on screen observations. (Cf. Appendix~\ref{appendix:mobile})

\vpara{VAB-WebArena-Lite~\cite{zhou2023webarena}.}
Web browsing is an ideal testbed for evaluating LMMs as GUI agents. Previous works~\cite{shi2017world,liu2018reinforcement,deng2024mind2web,yao2022webshop} mainly focus on offline evaluation. We adopt WebArena~\cite{zhou2023webarena}, a benchmark for text-based web GUI agents with \num{812} tasks across \num{5} websites. LMMs perform tasks based on user instructions, such as finding and summarizing customer reviews on \texttt{OneStopShop}. We use HTML SoM~\cite{koh2024visualwebarena} to annotate operable HTML elements, enabling LMMs to generate actions via \texttt{playwright}. WebArena-Lite is a subset of \num{165} tasks, refined and adapted for multimodal evaluation, removing cross-website tasks and fixing implausible conditions. (Cf. Appendix~\ref{appendix:webarena})

\subsection{Visual Design Agent}
Visual design tasks demand a nuanced understanding of visual signals, 
which text-only LLMs cannot handle with any easy augmentation, 
unlike embodied or GUI agent tasks that can rely on external object detectors~\cite{llm-planner} or textual representations like accessibility trees~\cite{xie2024osworld}.

\vpara{\sm-CSS.}
We create a new task to evaluate LMMs on web frontend design, focusing on CSS style adjustments.
Fixing CSS styles is a labor-intensive task that often requires engineers to iteratively adjust an element through trial and error.
Such a task inherently entails fine-grained visual grounding and reasoning across a series of rendering outcomes resulting from iterative CSS edits.
In \sm-CSS, the agent iteratively edits the CSS style using provided tools until it thinks the rendering matches a given target design.
We adopt \textit{success rate (SR)} as the metric, which evaluates whether the final rendering matches the target design. (Cf. Appendix~\ref{appendix:css})

\section{Methodology for \sm Data Collection}
\begin{figure}[t]
    \centering
    \includegraphics[width=\linewidth]{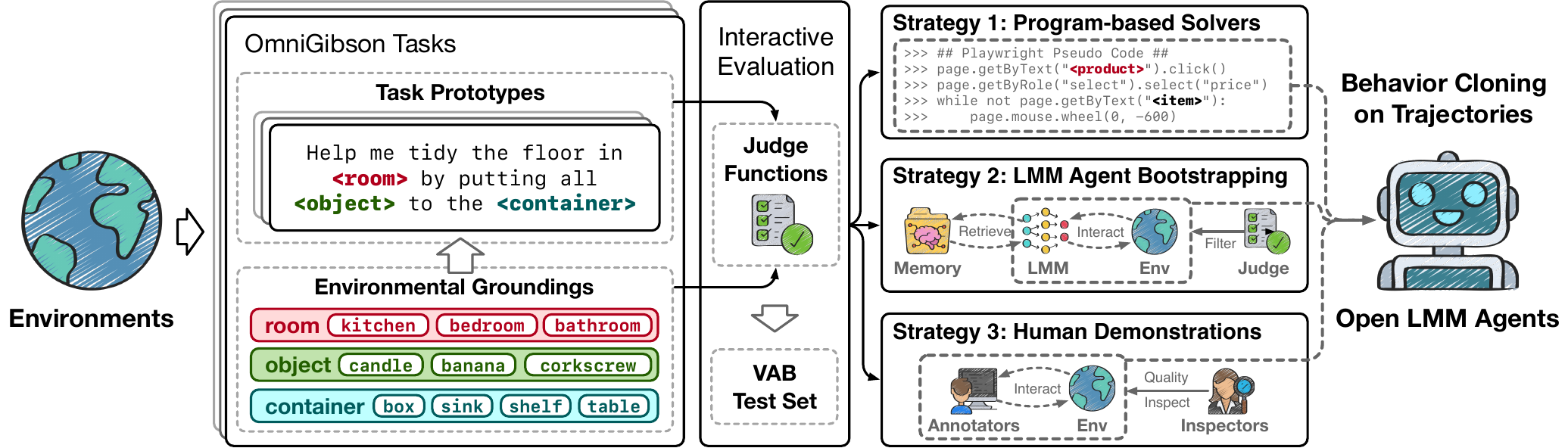}
    \caption{Data collection process in \sm. We follow a principled two-stage paradigm to collect task instances and then adopt various methods to further collect training trajectories for them.}
    \label{fig:method}
    \vspace{-5mm}
\end{figure}

For agent tasks, it is known to be very challenging to design practical and verifiable task instances; let alone creating high-quality training trajectories on top of them later.
In constructing \sm, we not only aim to deliver a high-quality agent benchmark but also endeavor to develop a systematic methodology for the problem of \problem data curation.
For task instance collection, we follow a two-stage paradigm (\textit{prototyping} and \textit{instantiation}) for each new task instance to ensure data quality and executability.
Additionally, we harness a suite of hybrid strategies to collect training trajectories that can be used to tune open LMMs into better visual foundation agents.
Our rigorous data collection process in \sm is crucial for presenting a high-quality resource for LMM-based agents (Figure~\ref{fig:method}).
The statistics of different tasks in \sm are shown in Table~\ref{tab:statistics}.

\subsection{Task Instance Collection: Prototyping and Instantiation}
Curating meaningful and testable task instances for LMM agent tasks can be difficult.
On one hand, they should be diverse and useful to cover real-world applications.
On the other hand, they should be grounded to environments carefully to ensure feasibility and practicality.
As a result, we collect all our task instances in a two-stage paradigm:

\begin{itemize}[leftmargin=1.5em,itemsep=0pt,parsep=0.2em,topsep=0.0em,partopsep=0.0em]
    \item \textbf{Prototyping:} We gather many task prototypes representing high-level goals based on the functionality provided by the environment. Related items are temporarily set to placeholders.
    \item \textbf{Instantiation:} Task prototypes are grounded to concrete items and conditions collected from the environment. Judging functions are thereby set up by instantiated tasks. Instructions are then rephrased by LLMs to enhance expression diversity.
\end{itemize}

For \sm-OmniGibson, a prototype is a general household activity, such as recycling office papers.
We source these prototypes either by sampling from Behavior-1K or by annotating them ourselves.
Instantiating a prototype involves grounding it in a specific scene  (e.g., specific rooms with office papers and recycling bins) generated in OmniGibson.
To increase task diversity, we instantiate each prototype with multiple random scenes and various initializations of object positions in the room.
In total, we collect \num{992} instances using \num{89} prototypes.
We sample \num{181} out of them as our test set.

For \sm-Minecraft, we target high-level task prototypes related to object collecting and then instantiate them with game configurations using different world seeds or spawn points. 
We manually check to ensure that each high-level goal is achievable within its configuration. 
In total, we collect \num{628} task instances using high-quality prototypes defined by us, with \num{116} instances designated as the test set. 
Additionally, we sample \num{132} task prototypes from JARVIS-1, resulting in \num{596} task instances that could be leveraged to collect our training trajectories later.

For \sm-Mobile, we first select \num{8} typical Android applications, from system services to third-party applications (e.g., Maps, Music, etc.) that could be evaluated offline.
We come up with \num{119} test instructions for them and prepare valid groundings in the AVD snapshot (e.g., an MP3 file to play in the Music APP).
For the training task construction, we filter \num{18} commonly used APPs and summarize their major functions to around \num{70} task prototypes.

For WebArena-Lite, we filtered and cleaned \num{165} test samples from the original WebArena dataset and collected new task instances for web applications to use in training trajectory collection.
Specifically, we summarize each website's basic functions and valid items for synthetic queries, created \num{40} task prototypes, and fill them with valid and invalid items (e.g., \texttt{product categories}, \texttt{prices}) to generate specific instructions, resulting in \num{1186} training task instances.

For \sm-CSS, a task prototype simply corresponds to one possible corruption of a CSS rule such as adding or altering a CSS property. 
To instantiate a task for a specific website, we randomly select a corruption that results in noticeable visual changes, determined by an SSIM~\cite{ssim} score below \num{0.8}.\footnote{This is an empirical choice based on our own experience.} 
In addition, we manually annotate each instance with a natural language description of the difference between the two images as an additional clue to the agent.
In total, we collect \num{1210} instances and use \num{165} to form the test set. 


\begin{table}[t]
\centering
\footnotesize
\caption{Recommendation levels for 3 strategies used in curating \sm's agent-tuning trajectory data on different dimensions. (Cf. Section~\ref{subsection:training} for detailed explanation on each dimension)}
\renewcommand\tabcolsep{2pt}
\resizebox{\columnwidth}{!}{
\begin{tabular}{@{}lccccc@{}}
\toprule
Strategy                & Avg. Cost          & Adaptability       & Versatility          & Flexibility        & Adoption                           \\ \midrule
Program-based Solvers   & \ystar\ystar\ystar & \ystar\ystar       & \ystar             & \ystar             & \sm-OmniGibson, \sm-WebArena-Lite  \\
LMM Agent Bootstrapping & \ystar             & \ystar\ystar\ystar & \ystar\ystar       & \ystar\ystar\ystar & \sm-Minecraft, \sm-Mobile, \sm-CSS \\
Human Demonstrations    & \ystar\ystar       & \ystar             & \ystar\ystar\ystar & \ystar\ystar       & \sm-Mobile                         \\ \bottomrule

\end{tabular}
}
\vspace{-2mm}
\label{tab:strategy_cmp}
\end{table}

\subsection{Training Trajectory Collection: 3-Leveled Strategies}
\label{subsection:training}
Recently, there has been a rise in benchmarks for interactively evaluating LLM or LMM agents~\cite{liu2023agentbench,zhou2023webarena,koh2024visualwebarena,xie2024osworld}.
Despite showcasing the substantial potential of LLM and LMM as agents, they usually only provide the test set and thus fail to facilitate the improving of open LLMs and LMMs on agent tasks.
In light of the challenge, in \sm we are devoted to offering a preliminary behavior cloning setup for training open LMM agents.

Imitation learning, especially the behavior cloning (BC)~\cite{nakano2021webgpt,zeng2023agenttuning} method, has been demonstrated as effective for building LLM agents from scratch.
However, a significant challenge lies in curating high-quality BC trajectories in large quantities, where the best strategy could be likely environment-depended.
In \sm, we systematically summarize our trajectory collecting into 3-leveled strategies:

\begin{enumerate}[leftmargin=1.5em,itemsep=0pt,parsep=0.2em,topsep=0.0em,partopsep=0.0em]
    \item \textbf{Program-based Solvers:} Trajectories are collected by prototype-specific programs written by human experts (e.g., \texttt{Playwright} scripts for automating web browsing tasks).
    \item \textbf{LMM Agent Bootstrapping:} Trajectories are collected by prompted LMM agents (e.g., \texttt{gpt-4o}), with optional memory augmentations~\cite{wang2023jarvis} to enhance performance. For instance, in Minecraft we allow agent to access memories for solving easier sub-goals (e.g., how to collect a stick) when constructing trajectories for more complex goals (e.g., how to collect a hammer).
    \item \textbf{Human Demonstrations:} Trajectories are annotated by human experts. It is necessary for scenarios where humans are indispensable (e.g., mobile apps require logged-in human accounts).
\end{enumerate}

These strategies are quite different from each other and present their own unique advantages in certain environments.
We summarize their recommendation levels on 4 dimensions (Cf. Table~\ref{tab:strategy_cmp}):

\begin{itemize}[leftmargin=1.5em,itemsep=0pt,parsep=0.2em,topsep=0.0em,partopsep=0.0em]
    \item \textbf{Average Cost:} The most important dimension. Program-based solvers are the cheapest for massive production. Human demonstrations are of medium average cost, especially in large quantities as annotators become more proficient over time. Bootstrapping is currently the most expensive, since it is so far necessary to harness proprietary LMM APIs for building visual agents, which are expensive but still suffer from high failure rates in many tasks. However, as open LMMs become stronger for visual agent tasks, the cost will be substantially reduced.
    \item \textbf{Adaptability:} It indicates how easy we can implement a strategy to an environment. Bootstrapping can be easily adapted to environments given good system prompts. Program-based solvers take some days for researchers to implement. Humans need detailed annotation manuals and sufficient time for training, and could be helpless due to poor environmental accessibility (e.g., hardware).
    \item \textbf{Versatility:} It refers to how versatile tasks a strategy could deal with. Well-trained annotators could accomplish almost all tasks, while LMM agents usually fail to handle difficult ones. Program-based solvers can only tackle given prototypes. Therefore, for situations where diverse instructions are needed (e.g., 18 apps involved in \sm-Mobile), versatility is a first concern.
    \item \textbf{Flexibility:} It denotes the trial and error process in the solution trajectories, which is crucial for agent applications where there could exist ideal but impractical single-step solutions. While LMM bootstrapping naturally presents the process, it is unlikely for program-based solvers to act so. For humans, trial and error in annotation is usually discouraged due to quality control.
\end{itemize}

Considering all mentioned dimensions and their trade-offs, we adopt a hybrid set of strategies for each of the 5 environments in \sm as shown in Table~\ref{tab:strategy_cmp}:

For \sm-OmniGibson, we adopt the program-based solvers focusing on the cost and adaptability. OmniGibson has no friendly interface for humans to operate on, and requires high-end laptops with GPUs supporting ray tracing and large main memory (> \num{10} GB) to run.
Thus it is unlikely for us to find a large number of qualified annotators to label for OmniGibson.
LMM agent bootstrapping is fine but uneconomical, as the task usually takes more steps than others (i.e., up to \num{100}).
Program-based solvers, instead, are suitable for collecting massive high-quality trajectories in OmniGibson.

For \sm-Minecraft, we adopt LMM agent bootstrapping considering adaptability.
Minecraft requires some flexible explorations (as environments are generated randomly), which is beyond the scope of program-based solvers.
Humans need to be well-trained for some time on playing Minecraft before becoming qualified annotators.
Since previous work has explored the usage of memory augmentation~\cite{wang2023jarvis} for improving LMM agents in Minecraft, it becomes practical to leverage the bootstrapping strategy by LMM APIs such as \texttt{gpt-4o} for creating training trajectories.

For \sm-Mobile, we primarily adopt human demonstrations, accompanied with some LMM Agent Bootstrapping considering the versatility and flexibility.
As android XMLs are less legible and operable than HTMLs on web with existing automation tools, program-based solvers are not applicable.
Additionally, for many apps require login and internet connection, human demonstration is the best solution.
LMM agent bootstrapping is employed in some offline APPs such as system settings to enhance trajectory flexibility.

For \sm-WebArena-Lite, we adopt program-based solvers due to cost and adaptability.
On the one hand, there have been a mature web automation tool \texttt{Playwright} that supports Python.
On the other hand, although WebArena~\cite{zhou2023webarena} is adopting some mirror websites for their real-world counterparts, their interfaces could be vastly different (e.g., OpenStreetMap in WebArena vs. Google Maps in real-world).
Consequently, human annotators struggle to label demonstrations on these websites efficiently in our preliminary trials.
For LMM agents, they tend to perform too poorly under mere prompting on WebArena (with success rate less than \num{20}\%) for efficient trajectory construction.

For \sm-CSS, we adopt LMM agent bootstrapping, mostly owing to concerns on flexibility.
A critical challenge for the agent in debugging CSS styles is to iteratively adjust the CSS rules through a trial and error process,
which can be flexibly achieved using the LMM agent bootstrapping scheme.
In particular, we first use \texttt{gpt-4o} to collect trajectories that finally resolve the CSS issue.
However, \texttt{gpt-4o} can only achieve a success rate lower than \num{40}\%.
To collect additional trajectories, we hint the agent with the target CSS rule to edit, after \num{5} steps of trials, on tasks where the agent initially fails.

\begin{table}[t]
\centering
\footnotesize
\caption{Comparison between \sm and related benchmarks. \sm is the first comprehensive multi-domain agent benchmark offering interactive environments, supporting multimodal agent evaluation, and providing a large and diverse set of training trajectories for visual agent tuning. ``\#Test Ins.'' refers to the number of test instances; ``\#Train Traj.'' refers to the number of training trajectories for SFT, ``RL'' means no training trajectory is available and only a reinforcement learning setup is provided; ``\#Avg. Turns'' refers to the average number of turns per training trajectory.}

\vspace{1mm}
\renewcommand\tabcolsep{2pt}
\renewcommand\arraystretch{1}
\resizebox{\columnwidth}{!}{
\begin{tabular}{@{}lccccccc@{}}
\toprule
  &  Category & \#Env. & \#Test Ins. & \#Train Traj.  & \#Avg. Turns & Multimodal  & Interactive Env. \\
\midrule
ALFWorld~\cite{shridhar2020alfworld}        & Household         & \num{1} & \num{134}   & \num{6374}      & \num{7.54} & \redx       & \greencheck   \\
Alfred~\cite{shridhar2020alfred}            & Household         & \num{1} & \num{1529}  & \num{6574}     & \num{7.51} & \greencheck & \greencheck  \\
Behavior-1K~\cite{li2023behavior}            & Household         & \num{1} & \num{1000}  & RL     & - & \greencheck & \greencheck  \\
MineDojo~\cite{fan2022minedojo}             & Game              & \num{1} & \num{3141}  & RL        & - & \greencheck & \greencheck  \\
SmartPlay~\cite{wu2023smartplay}            & Game              & \num{6} & \num{20}    & -     & - & \redx       & \greencheck   \\
Mind2Web~\cite{deng2024mind2web}            & Web               & \num{1} & \num{1341}  & \num{1009}      & \num{7.71} & \greencheck & \redx        \\
WebArena~\cite{zhou2023webarena}            & Web               & \num{1} & \num{812}   & -     & - & \greencheck & \greencheck  \\
VisualWebArena~\cite{koh2024visualwebarena} & Web               & \num{1} & \num{910}   & -     & - & \greencheck & \greencheck  \\
META-GUI~\cite{sun2022meta}                 & Mobile            & \num{1} & \num{483}   & \num{3692}       & \num{7.64} & \greencheck & \redx     \\
OSWorld~\cite{xie2024osworld}               & Desktop           & \num{1} & \num{369}   & -     & - & \greencheck & \greencheck  \\
OmniACT~\cite{kapoor2024omniact}            & Desktop \& Web    & \num{2} & \num{9802}  & -     & - & \greencheck & \redx       \\
AgentBench~\cite{liu2023agentbench}         & Multi-domain     & \num{8} & \num{1091}  & -     & - & \redx       & \greencheck  \\ \midrule
\model                                         & Multi-domain     & \num{5} & \num{746}   & \num{4482}      & \num{11.22} & \greencheck & \greencheck  \\
\bottomrule
\end{tabular}
}
\label{table:related}
\vspace{-5mm}
\end{table}

\begin{table}[t]
    \centering
    \footnotesize
    \caption{Statistics of all datasets in \sm. }
    \renewcommand\tabcolsep{4pt}
    \renewcommand\arraystretch{0.9}
    \resizebox{\textwidth}{!}{
    \begin{tabular}{@{}lcccccc@{}}
        \toprule
        \textbf{} & \sm-OmniGibson & \sm-Minecraft & \sm-Mobile & \sm-WebArena-Lite & \sm-CSS \\
        \midrule
        \#Action Space & \num{20} & \num{6} & \num{7} & \num{12} & \num{4} \\
        \#Test Instance & \num{181} & \num{116} & \num{119} & \num{165} & \num{165} \\
        \#Train Trajectory & \num{872} & \num{382} & \num{1213} & \num{1186} & \num{829} \\
        \#Train Step & \num{20153} & \num{5197} & \num{10175} & \num{9522} & \num{5250} \\
        \#Max Round Limit & \num{100} & \num{100} & \num{25} & \num{20} & \num{10}\\
        \bottomrule
    \end{tabular}
    }
    \label{tab:statistics}
    \vspace{-6mm}
\end{table}

\section{Baseline Experiment}

Using \sm, we evaluate a comprehensive array of proprietary LMMs with prompting and also some selected open LMMs with fine-tuning to serve as LMM-as-Visual-Agent baselines.
We also dive into several insights we encounter during the testing of existing LMMs, which unveil the critical challenges and future research directions for the development of LMM agents.

\subsection{Setup}

\vpara{Baselines.}
We evaluate on both proprietary LMM APIs and selected open LMMs.
For proprietary LMMs, we include models from OpenAI GPT~\cite{GPT-4o,GPT-4-vision-preview, GPT-4o-mini}, Anthropic Claude~\cite{Claude-3}, Google Gemini~\cite{reid2024gemini,team2023gemini}, and Qwen-VL-Max~\cite{bai2023qwen}.
For open LMMs, we select eight state-of-the-art models as representative fine-tuning baselines in \sm: InternVL-2~\cite{chen2024far}, GLM-4V~\cite{glm2024chatglm}, CogVLM2~\cite{wang2023cogvlm}, CogAgent~\cite{hong2023cogagent}, CogVLM~\cite{wang2023cogvlm}, LLaVA-NeXT~\cite{liu2024llavanext}, LLaVA-1.5~\cite{liu2024improved}, Qwen-VL~\cite{bai2023qwen}.

\vpara{Prompting.}
We format \problem as two roles (i.e., \texttt{user} and \texttt{assistant}) interacting in multiple rounds.
The task description, action spaces, few-shot demonstrations, and important notices for each environment are formatted as the \texttt{system} prompt at the beginning of the conversation.
Task instruction is given in the first \texttt{user} round. 
Environmental observations and feedback are passed via \texttt{user} in later rounds.
Considering current LMM APIs' poorer support of multi-image and outrageous cost when interaction rounds soar up, in Embodied and GUI agents we only offer the vision input of the latest \texttt{user} round (following~\cite{koh2024visualwebarena}) while reserving history text contents.
An exception is the CSS agent in Visual Design. In this case, comparing differences in visual inputs is essential, and the interaction rounds are typically fewer than \num{10}. 
Therefore, we retain all image inputs in the conversation history for this task.

\begin{table}[t]
\footnotesize
\renewcommand\tabcolsep{1.5pt}
\renewcommand\arraystretch{1.2}
\caption{Main results on \model. The metric reported is success rate (SR), which measures the proportion of successful tasks in all evaluated tasks. Open LMMs are evaluated using multitask fine-tuning rather than direct prompting, as they were unable to effectively follow system prompts from \sm in our preliminary trials. For \# Params of open LMMs, we report the sizes of their language and vision part respectively.}
\vspace{2mm}
\resizebox{\columnwidth}{!}{
\begin{tabular}{@{}clc|c|ccccc@{}}
\toprule
\multirow{2}{*}{Type}                                                                     & \multirow{2}{*}{Model}                                    & \multirow{2}{*}{\# Params} & \multirow{2}{*}{AVG} & \multicolumn{2}{c}{Embodied}  & \multicolumn{2}{c}{GUI}       & Visual Design \\ \cmidrule(l){5-6} \cmidrule(l){7-8} \cmidrule(l){9-9}
                                                                                          &                                                           &                            &                      & OmniGibson    & Minecraft     & Mobile        & WebArena-Lite & CSS           \\ \midrule
\multirow{9}{*}{\begin{tabular}[c]{@{}c@{}}Proprietary\\ LMMs\\ (Prompting)\end{tabular}} & \texttt{gpt-4o-2024-05-13}~\cite{GPT-4o}                             & N/A                        & \textbf{\num{36.2}}        & \textbf{\num{41.4}} & {\ul \num{55.2}}    & \textbf{\num{31.9}} & \num{18.2}          & \textbf{\num{34.5}} \\
                                                                                          & \texttt{gpt-4-vision-preview}~\cite{GPT-4-vision-preview} & N/A                        & {\ul \num{31.7}}           & {\ul \num{36.5}}    & \num{47.4}          & \num{26.9}          & {\ul \num{18.8}}    & {\ul \num{29.1}}    \\
                                                                                          & \texttt{gpt-4-turbo-0409}~\cite{GPT-4-vision-preview}     & N/A                        & \num{29.9}                 & \num{26.5}          & \num{50.0}          & \num{26.9}          & \num{18.2}          & \num{27.9}          \\
                                                                                          & \texttt{claude-3.5-sonnet}~\cite{Claude-3.5-Sonnet}       & N/A                        & \num{26.9}                 & \num{24.3}          & \textbf{\num{56.0}} & {\ul \num{31.1}}    & \num{7.2}           & \num{15.8}          \\
                                                                                          & \texttt{claude-3-opus}~\cite{Claude-3}                    & N/A                        & \num{21.9}                 & \num{14.9}          & \num{51.7}          & \num{15.1}          & \num{7.9}           & \num{20.0}          \\
                                                                                          & \texttt{gpt-4o-mini-2024-07-18}~\cite{GPT-4o-mini}        & N/A                        & \num{20.5}                 & \num{12.2}          & \num{30.2}          & \num{22.7}          & \textbf{\num{20.6}} & \num{17.0}          \\
                                                                                          & \texttt{gemini-1.5-pro}~\cite{reid2024gemini}             & N/A                        & \num{19.8}                 & \num{22.1}          & \num{41.4}          & \num{16.8}          & \num{7.9}           & \num{10.9}          \\
                                                                                          & \texttt{gemini-1.0-pro}~\cite{team2023gemini}             & N/A                        & \num{6.3}                  & \num{4.4}           & \num{11.2}          & \num{11.8}          & \num{4.2}           & \num{0.0}           \\
                                                                                          & \texttt{qwen-vl-max}~\cite{bai2023qwen}                   & N/A                        & \num{2.7}                  & \num{0.0}           & \num{6.0}           & \num{2.5}           & \num{3.0}           & \num{1.8}           \\ \midrule
\multirow{8}{*}{\begin{tabular}[c]{@{}c@{}}Open LMMs\\ (Fine-tuning)\end{tabular}}        & \texttt{InternVL-2}~\cite{chen2024far}                    & \num{7}B + \num{0.3}B                  & \num{16.0}                 & \num{16.0}          & \num{28.4}          & \num{3.4}           & \num{7.9}           & \num{24.2}          \\
                                                                                          & \texttt{GLM-4V}~\cite{glm2024chatglm,wang2023cogvlm}      & \num{9}B + \num{4}B                    & \num{12.0}                 & \num{8.8}           & \num{19.8}          & \num{2.5}           & \num{5.5}           & \num{23.6}          \\
                                                                                          & \texttt{LLaVA-NeXT}~\cite{liu2024llavanext}               & \num{8}B + \num{0.3}B                  & \num{10.5}                 & \num{3.3}           & \num{23.3}          & \num{3.4}           & \num{4.2}           & \num{18.2}          \\
                                                                                          & \texttt{CogVLM2}~\cite{wang2023cogvlm}                    & \num{8}B + \num{12}B                   & \num{10.3}                 & \num{3.3}           & \num{25.9}          & \num{1.7}           & \num{3.0}           & \num{17.6}          \\
                                                                                          & \texttt{CogAgent}~\cite{hong2023cogagent}                 & \num{7}B + \num{11}B                   & \num{8.9}                  & \num{6.6}           & \num{20.7}          & \num{2.5}           & \num{0.6}           & \num{13.9}          \\
                                                                                          & \texttt{CogVLM}~\cite{wang2023cogvlm}                     & \num{7}B + \num{10}B                   & \num{8.4}                  & \num{3.3}           & \num{19.8}          & \num{4.2}           & \num{4.2}           & \num{10.3}          \\
                                                                                          & \texttt{LLaVA-1.5}~\cite{liu2024improved}                 & \num{13}B + \num{1}B                   & \num{7.7}                  & \num{1.7}           & \num{25.9}          & \num{0.8}           & \num{2.4}           & \num{7.9}           \\
                                                                                          & \texttt{Qwen-VL}~\cite{bai2023qwen}                       & \num{7}B + \num{2}B                    & \num{5.7}                  & \num{1.7}           & \num{18.1}          & \num{1.7}           & \num{2.4}           & \num{4.8}           \\ \bottomrule
\end{tabular}
\label{tab:main}
}
\vspace{-5mm}
\end{table}

\vpara{Training for Open LMMs.}
We generally follow the prompting format of proprietary LMM APIs in each environment to organize our training trajectories, and make several minor modifications.
In the system prompt we remove the few-shot demonstrations as we would fine-tune models.
In addition, during fine-tuning, since open LMMs perform poorly on multi-image input (especially for CogVLM and CogAgent, whose expert architecture disallows simple implementation of multi-image input), we only use the vision input of the latest \texttt{user} turn, and concatenate histories together using role tokens (i.e., ``\texttt{<|user|>}'') and linebreaks.
For CSS agent where multi-image input is necessary, we concatenate history images vertically into one as the input.
To benchmark the potential of LMMs to serve as visual foundation agents, we conduct multitask fine-tuning over the dataset aggregation of all environments.
To optimize performance, all LMMs undergo full-parameter fine-tuning, with a batch size of \num{64} and \num{5}k training steps.
Other hyperparameters are configured using the default ones provided by the model's original repository or the third-party's integrated training framework.
For data composition, we uniformly combine all training samples except for \sm-CSS, which we duplicate an additional \num{2} times as the preliminary experiments show that the task requires more extensive training for open LMMs to adapt to the screenshot concatenation format.

\subsection{Main Results}
Table~\ref{tab:main} show the main results on \sm, including both prompting proprietary LMMs and fine-tuned open LMMs.
We have several important observations on the status quo of \problem.

\vpara{\sm is challenging for existing LMMs.}
We observe that existing LMMs face significant challenges when evaluated on \sm. 
The majority of proprietary LMMs, with mere prompting, achieve an overall success rate above \num{20}\%, demonstrating their multimodal understanding and reasoning abilities. 
The most capable LMM, \texttt{gpt-4o}, achieves an overall success rate of \num{36.2}\%. However, these performances are still far from satisfactory and not yet qualified for direct deployment.
Notably, despite its superiority on existing benchmarks, \texttt{claude-3.5-sonnet} still falls significantly behind \texttt{gpt-4o}. 
Additionally, we present the first systematic evaluation of \texttt{gpt-4o-mini} on agent tasks, which reveals that its performance is considerably inferior to \texttt{gpt-4o} but comparable to \texttt{claude-3-opus} and \texttt{gemini-1.5-pro}.

\vpara{Trajectory SFT can improve LMM agents.}
For open LMMs, we find they can rarely follow the system prompt's instruction without fine-tuning in preliminary trials, resulting in \num{0}\% success rates.
After training on \sm, open LMMs present significant improvements.
The strongest one, \texttt{InternVL-2}, even outperforms \texttt{gemini-1.0-pro} on all evaluated environments and \texttt{claude-3-opus} on CSS agent task.
These results suggest that learning from trajectories would be a promising direction for us to build visual foundation agents.

\vpara{Gaps between top proprietary and open LMMs are huge but likely to be narrowed.}
Despite the improvement from trajectory SFT, the gap between proprietary and tested open LMMs is much wider than expected.
While many of them have claimed competitive performance to \texttt{gpt-4-vision-preview} on traditional vision benchmarks such as image captioning, VQA, and so on, their fundamental ability to serve as practical visual foundation agents is far from comparable even after fine-tuning on \sm datasets.
It also demonstrates that \sm could serve as an ideal testbed for benchmarking the practical performance of LMMs.
With larger backbone LLMs (which are insufficiently tested in this work due to limitations of our computing resources) and more high-quality trajectory data, it is likely that open LMMs will be comparable or even outperform more proprietary LMMs.

\section{Analysis} \label{sec:analysis}
Multimodal agent tasks encompass two critical challenges: \textit{visual grounding} and \textit{planning}.
We conduct fine-grained analyses to gain deeper insights into performance in these two aspects and offer valuable perspectives for the future development of visual foundation agents based on LMMs.

\subsection{Visual Grounding Analysis}
Visual grounding refers to the ability to associate language concepts with content in visual perception~\cite{fukui2016multimodal,zheng2024gpt4vision}, which is crucial for \problem.
We look into 3 typical design choices in \sm related to visual grounding to show its current status and challenges.

\begin{figure}[ht]
    \centering
    \begin{minipage}{0.5\textwidth}
        \centering
        \includegraphics[width=\linewidth]{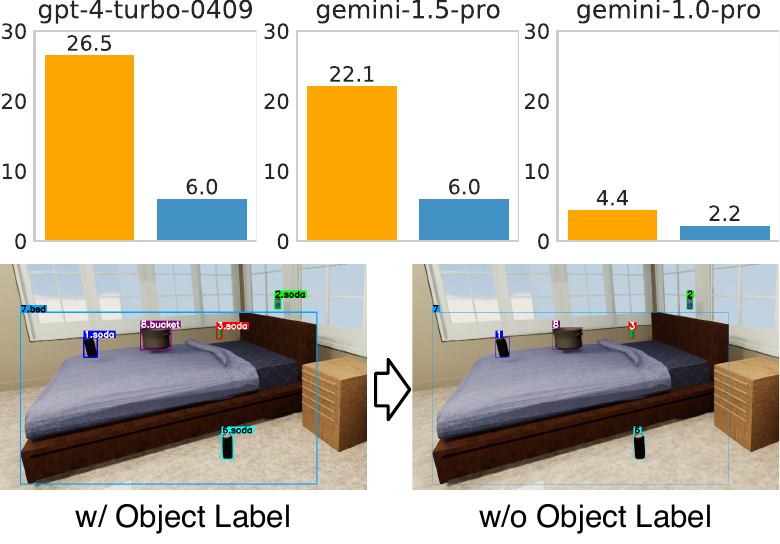}
        \caption{Comparing w/ and w/o Object Labels.}
        \label{fig:wo_tag}
    \end{minipage}\hfill
    \begin{minipage}{0.4\textwidth}
        \centering
        \includegraphics[width=\linewidth]{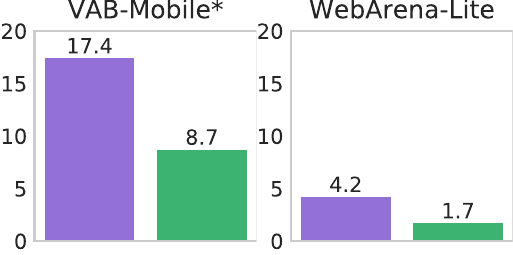}
        \caption{Comparing SoM and REC in GUI agent tasks, trained on CogVLM2. \sm-Mobile$^*$ here is an earlier version different from the one in Table~\ref{tab:main}.}
        \label{fig:gui_som}
    \end{minipage}
\end{figure}

\vpara{The use of object labels in embodied environment.}
Despite the strong image caption and object recognizing ability of LMMs, they do not seem to play well in the context of an embodied agent task.
In \sm-OmniGibson, we compare the \problem performance with and without object labels annotated in the vision input.
The result shows that LMM agents significantly underperform without object labels.
It indicates that notwithstanding LMMs' strong performance on downstream benchmarks, they can still struggle in the same task in the context of \problem.

\vpara{The use of Set-of-Marks (SoM) in GUI environment.}
For GUI tasks, we also augment the image input with SoM by default because it is difficult to elicit accurate bounding box coordinates from the LMM, which is essentially a referring expression comprehension (REC) task~\cite{qiao2020referring}.
With our training trajectories, we can evaluate whether LMMs can effectively perform visual grounding by directly outputting a bounding box without relying on external SoM signals.
Specifically, we fine-tune \texttt{CogVLM2} with and without SoM. The results in Figure~\ref{fig:gui_som} show that \texttt{CogVLM2} struggles to learn to directly output a bounding box, and SoM plays an instrumental role in visual grounding.

\begin{figure}[t]
    \centering
    \includegraphics[width=0.95\linewidth]{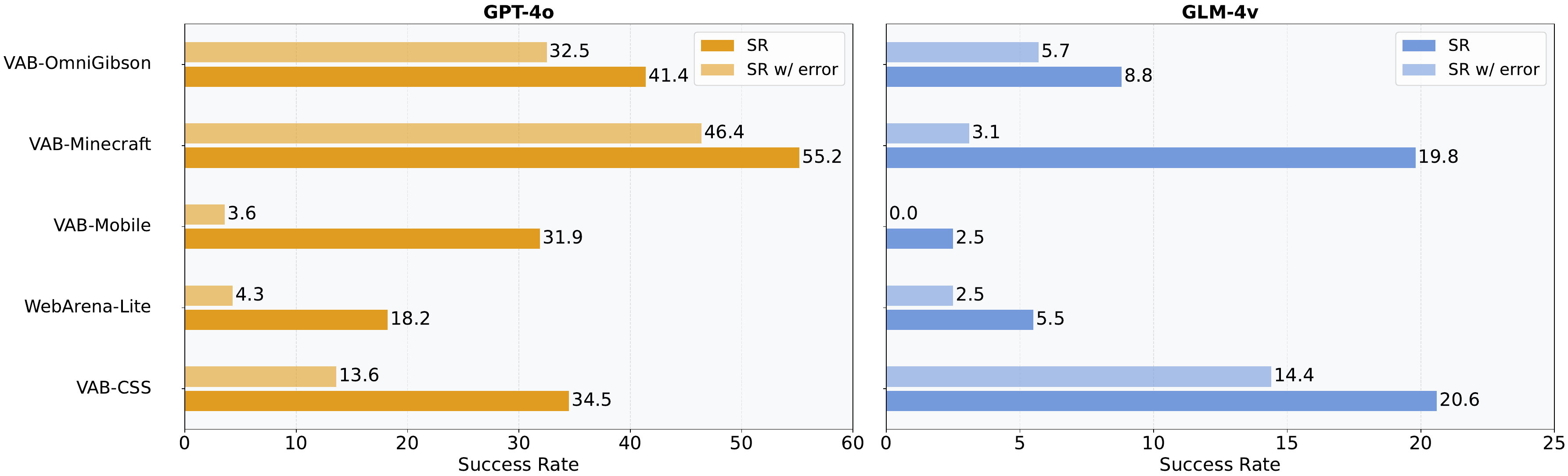}
    \caption{Comparison of overall success rates and success rates when incorrect actions are present in trajectories for various tasks.}
    \label{fig:error}
\end{figure}

\vpara{Visual difference grounding.}
Our new frontend design task provides us a unique opportunity to look into a specific type of visual grounding: \textit{visual difference grounding}.
Unlike traditional visual grounding with a single scene, which involves associating a natural language concept to a static region or object in the image, visual grounding in \sm-CSS requires the LMM to properly ground the ``layout difference'' (see our prompt in Appendix~\ref{appendix:css_prompt}) to the different areas of two images through comparison.
All our current results on \sm-CSS in Table~\ref{tab:main} are based on a relatively lenient setting. Instead of requiring the LMM to directly perceive the difference between two screenshots, we provide a natural language description that explicitly states the adjustments to make to match the two input images (see an example in Appendix~\ref{appendix:css_data}).
\begin{table}[t]
    \centering
    \footnotesize
    \caption{The performance of LMMs drops drastically on \sm-CSS when the natural language description is removed.}
    \begin{tabular}{lccc}
        \toprule
         & \texttt{gpt-4o-2024-05-13} & \texttt{gpt-4-turbo-0409} & \texttt{gpt-4-vision-preview}\\
        \midrule
        \textit{w/} NL  & \num{34.5}   & \num{27.9}   & \num{29.1}   \\
        \textit{w/o} NL  & \hspace{3px}\dscore{\num{24.2}}{\num{10.3}}   & \dscore{\num{1.9}}{\num{26.1}}   & \dscore{\num{2.4}}{\num{26.7}}   \\
        \bottomrule
    \end{tabular}
    \vspace{-0.3cm}
    \label{tab:vdg}
\end{table}
\begin{table}[t]
    \centering
    \footnotesize
    \caption{Directly generating an action leads to similar performance to ReAct.}
    \begin{tabular}{lcccc}
        \toprule
        Model & Prompting & \sm-Minecraft & \sm-Mobile & \sm-CSS\\
        \midrule
        \multirow[c]{2}{*}{\texttt{gpt-4o}} & \textit{w/} Thought  & \num{55.2}   & \num{30.4}   & \num{34.5}   \\
        & \textit{w/o} Thought  & \hspace{-5px}\dscore{\num{48.3}}{\num{6.9}}   & \hspace{-5px}\pscore{\num{31.9}}{\num{1.5}}   & \hspace{-5px}\pscore{\num{38.2}}{\num{3.7}}   \\
        \cmidrule{1-5}
        \multirow[c]{2}{*}{\texttt{claude-3.5-sonnet}} & \textit{w/} Thought  & \num{56.0}   & \num{29.0}   & \num{15.8}   \\
        & \textit{w/o} Thought  & \hspace{-5px}\dscore{\num{55.2}}{\num{0.8}}   & \hspace{-5px}\pscore{\num{31.1}}{\num{2.1}}   & \hspace{-5px}\pscore{\num{17.6}}{\num{1.8}}   \\
        \bottomrule
    \end{tabular}
    \vspace{-0.3cm}
    \label{tab:thought}
\end{table}

\subsection{Performance on Planning}
\vpara{The role of thought in ReAct.}
 ReAct~\cite{yao2022react} is one of the most commonly used frameworks for language agents. 
The central concept emphasizes the importance of integrating the agent's reasoning and actions by intertwining the output with both thought and action components.
However, in our study, we find that the thought step may not be essential. When using \texttt{gpt-4o} and \texttt{claude-3.5-sonnet} as the backbone of the agents, directly outputting an action field can yield comparable or even superior performance compared to using the ReAct framework (see Table~\ref{tab:thought}).

\vpara{Recovering from errors during planning.}
In real-world applications, agents must dynamically adjust their actions and plans based on environmental feedback. 
A crucial capability required for this is error recovery.
To understand error recovery capabilities in LMMs, we analyze two representative models: \texttt{gpt-4o}, the most powerful model currently available, and \texttt{glm-4v}, a prominent open LMM. 
Their performance, illustrated in Figure~\ref{fig:error}, reveals that \texttt{gpt-4o} exhibits robust error recovery across most tasks, with GUI tasks being an exception due to their often irreversible nature. 
Importantly, we find that incorporating error recovery scenarios in training data significantly enhances the performance of fine-tuned open LMMs, as evidenced by results from \sm-OmniGibson and \sm-CSS (Cf. Appendix~\ref{appendix:omnigibson-training} and Appendix~\ref{appendix:css-training} for details about error recovery of training trajectories).

\section{Related Work}

\vpara{LMM-as-Visual-Agent.}
In pre-LMM era, most visual agents are built with task specific training~\cite{shridhar2020alfred} and reinforcement learning~\cite{kempka2016vizdoom}. With the rapid development of LMMs~\cite{GPT-4o, reid2024gemini, GPT-4-vision-preview, bai2023qwen, Claude-3, team2023gemini, glm2024chatglm}, the study of LMM-based visual agents begins to thrive. Leveraging the general capabilities of LMMs, these visual agents have the potential to perform complex tasks in various scenarios, including embodied and game tasks~\cite{brohan2022rt, yang2023octopus, driess2023palm, tan2024cradleempoweringfoundationagents}, GUI interaction~\cite{zheng2024gpt4vision,zhou2023webarena,koh2024visualwebarena,xie2024osworld,kapoor2024omniact,yang2023appagent}, and visual design tasks~\cite{design2code,websight}.
However, these complex scenarios pose several challenges for LMM-based visual agents: basic visual understanding and grounding~\cite{zheng2024gpt4vision,yue2023mmmu}, vision-text information comprehension~\cite{Kil_2024_CVPR}, instruction following, and long-term planning ability~\cite{wu2023smartplay,liu2023agentbench}. Most general-purpose LMMs still lack strong zero-shot capabilities, leading to different application paradigms when deploying LMMs as visual agents. While prompting methods offer great convenience, they may not achieve satisfactory performance in many areas~\cite{zhou2023webarena,drouin2024workarena}. Consequently, task-specific training and alignment remain common practices in these applications~\cite{lai2024autowebglm}.
In response, \sm aims to establish a comprehensive benchmark for LMM-based visual agents, covering a wide range of typical applications. In the meantime, \sm seeks to provide an in-depth evaluation of both prompting and training approaches, ultimately fostering the development of LMM visual agents.

\vpara{Benchmarking LMM-based visual agents.}
With the rapid development of LMM agents and their impressive performance in various scenarios~\cite{xie2024osworld,kapoor2024omniact,yang2023appagent,yang2023octopus,design2code,mu2024embodiedgpt}, it has made the evaluation of LMM agent an urgent problem. In the GUI interaction domain, recent works have proposed static datasets~\cite{deng2024mind2web,rawles2024androidinthewild,sun2022meta} and interactive environments~\cite{zhou2023webarena,koh2024visualwebarena,xie2024osworld} to evaluate LMM agents in different applications, including web~\cite{zhou2023webarena, koh2024visualwebarena, deng2024mind2web}, mobile phone~\cite{rawles2024androidinthewild, sun2022meta}, and desktop~\cite{xie2024osworld}. In the embodied domain, previous works have proposed various game environments~\cite{guss2019minerl,fan2022minedojo} and household environments~\cite{li2023behavior}, but few works have explored benchmarking LMM agents on these environments. 
Most existing benchmarks are designed for relatively narrow domains and lack a comprehensive evaluation across different applications of LMM agents. Additionally, many benchmarks focus solely on the prompting evaluation of LMM agents. \sm aims to provide a training set for open-source foundation LMMs, offering a new perspective on benchmarking these models and advancing their applicability in diverse tasks.
\section{Conclusion}
We introduce a new benchmark for evaluating LMMs as visual agents. 
In addition, we also provide training trajectories essential for fine-tuning LMMs. 
However, open LMMs fine-tuned with our dataset still perform below the level of proprietary LMMs like \texttt{gpt-4o}. 
Behavior cloning with the offline collected trajectories is a vital first step, but future advancements will come from integrating reinforcement learning in diverse interactive environments.

\bibliographystyle{plain}
\bibliography{ref}
\clearpage

\part{Appendix} 
\parttoc 
\clearpage
\appendix
\section{\sm-OmniGibson}
\label{appendix:omnigibson}
In this section, we provide additional details about \sm-OmniGibson that are not covered in the main paper due to space limitations.

\subsection{Detailed Description}
Current household datasets or benchmarks are not originally designed for LMMs, making them less suitable for evaluating today's LMMs. Behavior-1K~\cite{li2023behavior} offers an action space focused on low-level physical control over the robot (e.g., joint angles), while Alfred~\cite{shridhar2020alfred} requires actions to output masks on images, which may not be practical for most LMMs. The ThreeDWorld Transport Challenge~\cite{gan2022threedworld} provides high-level action APIs, but the simulator environment is less realistic and the tasks may not fully challenge LMMs. The recent work Octopus~\cite{yang2023octopus} sets up household tasks for LMMs in the OmniGibson simulator. However, in this setting, vision input is less critical as the observed objects are also listed in text input for LMMs.

In order to set up a realistic and challenging benchmark for testing LMMs' embodied planning ability, we select the recent household simulator OmniGibson~\cite{li2023behavior} as the interactive environment, and build a pipeline for LMM to serve as a high-level planner on everyday household activities. An example of the task is shown in Fig.~\ref{fig:omnigibson_example}: The ego-centric image with annotated bounding boxes, high-level activity instruction and environment feedback are fed into the LMM, and it is tasked with reasoning over the current progress to decide on the next low-level action. It must interact with objects using the corresponding tags attached to the bounding boxes.

\vpara{Test Set.}
We select 45 activity instances from Behavior-1K~\cite{li2023behavior}, and manually adapt some of them to ensure these activities are solvable within our provided action space and suitable for evaluating current LMMs' embodied planning ability. We instantiate each activity in several scenes, resulting in a total of 181 test task instances. All the activity instructions are manually annotated by us.

\vpara{Training Set.}
\label{appendix:omnigibson-training}
We provide a set of successful trajectories using both rule-based solving and LMM bootstrapping. We newly design 47 activities, each instantiated in several different scenes with various initializations of object positions, resulting in a total of 901 task instances. To solve these tasks, we develop a rule-based solver that decomposes the long-horizon activities into subtasks and solves them sequentially. Running the rule-based solver on the 901 training task instances yields 785 successful trajectories. Then we manually add a type of error recovery process (agent fails to place an object into a closed container, and then opens the container) into these trajectories, aiming to enhance LMMs’ capability to rectify errors. Additionally, we select 464 training instances and utilize \texttt{gpt-4-vision-preview} to bootstrap 87 successful trajectories, resulting in a total of 872 training trajectories.

\vpara{Metrics.} We adopt task success rate as the metric of \sm-OmniGibson. In Behavior-1K~\cite{li2023behavior}, each activity is defined in the form of BEHAVIOR Domain Definition Language (BDDL)~\cite{srivastava2022behavior}, which describes the concrete initial and goal conditions of a specific activity. Only when all the goal conditions are met within the limit of 100 turns, the task is judged as successfully completed.

\subsection{Actions}
In \sm-OmniGibson, we provide the LMM agent with 20 low-level actions to interact with objects and navigate the household environment. The actions marked with an asterisk (*) are adapted from OmniGibson~\cite{li2023behavior}, while the others are newly defined and implemented by us. With these provided actions, the LMM agent is possible to solve all the testing instances.

\begin{itemize}[leftmargin=1.5em,itemsep=0pt,parsep=0.2em,topsep=0.0em,partopsep=0.0em]
\item \textbf{grasp}: Grasp a specific object into the robot's hand.
\item \textbf{move}: Move towards a specific object.
\item \textbf{move\_to\_room}: Move to a specific room in the house.
\item \textbf{turn\_left}: Turn the robot left 90 degrees.
\item \textbf{turn\_right}: Turn the robot right 90 degrees.
\item \textbf{raise\_camera}: Raise the camera of the robot to see higher objects.
\item \textbf{lower\_camera}: Lower the camera of the robot to see lower objects.
\item \textbf{put\_inside}: Place the object from the robot's hand inside another object.
\item \textbf{put\_on\_top}: Place the object from the robot's hand on top of another object.
\item \textbf{put\_under}: Place the object from the robot's hand under another object.
\item \textbf{put\_next\_to}: Place the object from the robot's hand next to another object.
\item \textbf{get\_fridge\_view}: Obtain the view inside a nearby fridge.
\item \textbf{cook}*: Cook a specific object.
\item \textbf{burn}*: Burn a specific object.
\item \textbf{freeze}*: Freeze a specific object.
\item \textbf{heat}*: Heat a specific object.
\item \textbf{open}*: Open a specific object.
\item \textbf{close}*: Close a specific object.
\item \textbf{toggle\_on}*: Turn on a specific object.
\item \textbf{toggle\_off}*: Turn off a specific object.
\end{itemize}

\subsection{Rule-based Solver for Training Trajectory Collection}

\vpara{BDDL task goals.}
Among activities of \sm-OmniGibson, each of the BDDL task goal can be decomposed into a sequence of subgoals (e.g., a specific door should be open, or a specific bottle should be on a specific countertop). All subgoals can be categorized into 2 types: identifying the state of a specific object, or the positional relationship between two objects.

\vpara{Method of rule-based solver.}
To achieve the BDDL task goal of a \sm-OmniGibson activity, the rule-based solver need to sequentially fulfill all the subgoals. For the first type of subgoal, the rule-based solver can navigate (\texttt{move\_to\_room}, \texttt{move}, \texttt{turn\_left}, \texttt{turn\_right}, \texttt{raise\_camera}, \texttt{lower\_camera}, \texttt{get\_fridge\_view}) to find the specific object; and then move towards it (\texttt{move}) and change its state (\texttt{cook}, \texttt{burn}, \texttt{freeze}, \texttt{heat}, \texttt{open}, \texttt{close}, \texttt{toggle\_on}, \texttt{toggle\_off}). For positional relationships, the solver should find and approach an object, grasp it (\texttt{grasp}), move to the other object, and finally complete the subgoal with \texttt{put\_inside}, \texttt{put\_on\_top}, \texttt{put\_under} or \texttt{put\_next\_to}.

\subsection{Prompt Example}
\begin{figure}[t]
    \centering
    \includegraphics[width=0.9\textwidth]{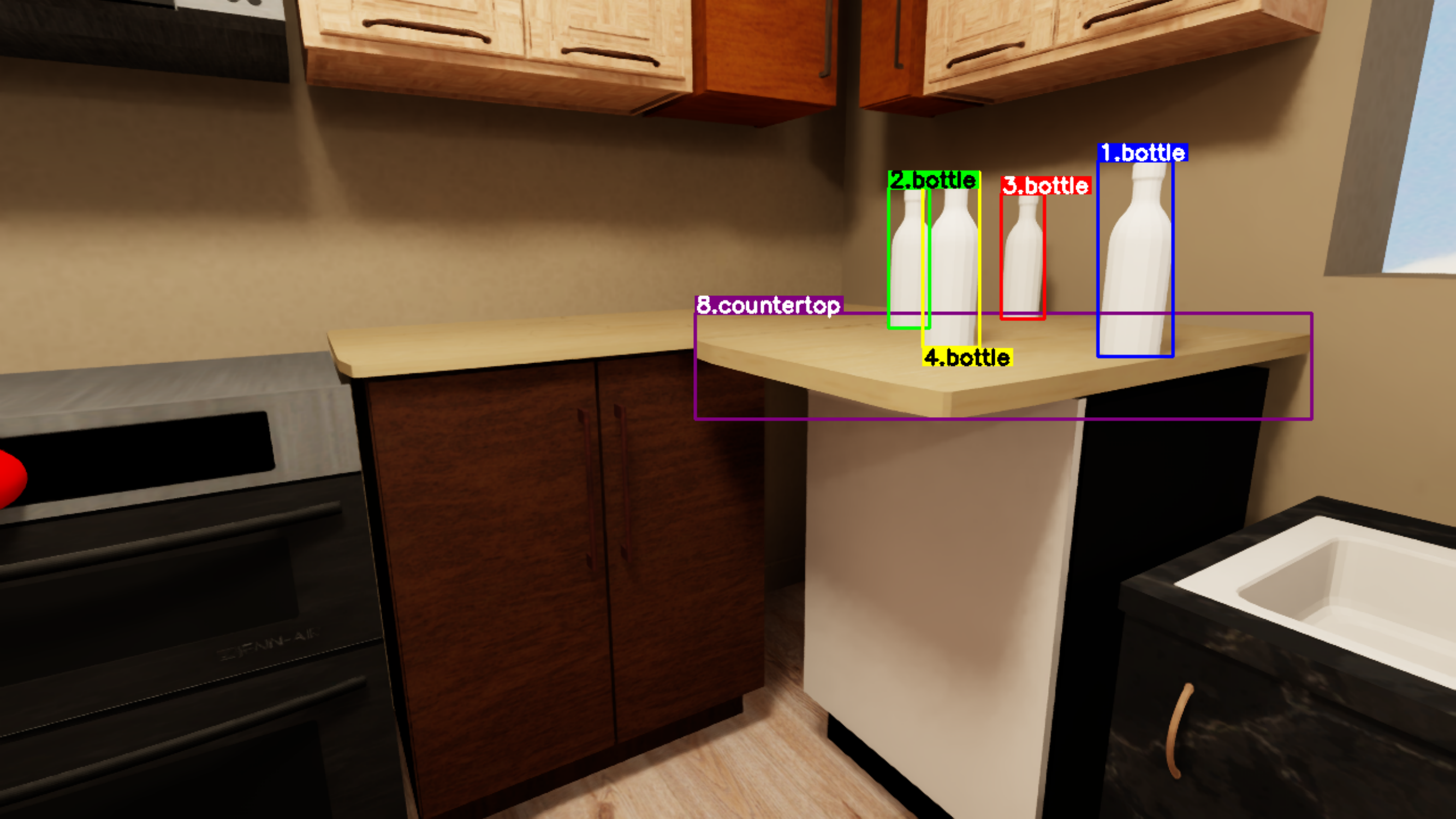}
    \caption{This is an example task of \sm-OmniGibson. The task asks the agent to bring all 4 bottles from the countertop into the fridge. The agent should grasp one bottle, navigate to find the fridge, open the fridge, put the grasped bottle into the fridge, and then repeat this process for the remaining bottles.}
    \label{fig:omnigibson_example}
\end{figure}
The system message that describes the detailed task information to the agent is shown as follows:
\lstset{
    backgroundcolor=\color[RGB]{245,245,244},
    breaklines=true,
    basicstyle=\ttfamily\small
}\begin{lstlisting}
# Setup
You are an intelligent agent exceling at solving household tasks. You are in a household environment given a task to finish.
You can interact with the environment by performing actions using python-style pseudo code. For each turn, please call exactly one predefined action.

# Valid Actions
## Predefined Action List:
```
def grasp(obj):
    '''Grasp the object in your hand.
    Args:
        :param obj: the digital identifier of the object to grasp.
    Returns:
        A string message of the environment feedback.
    '''
def move(obj):
    '''Move yourself towards the object.
    Args:
        :param obj: the digital identifier of the object to move towards.
    Returns:
        A string message of the environment feedback.
    '''
def move_to_room(room):
    '''Move yourself to a random position in the room.
    Args:
        :param room: the name of the room to move to.
    Returns:
        A string message of the environment feedback.
    '''
def turn_left():
    '''Turn the robot left 90 degrees.
    Returns:
        A string message of the environment feedback.
    '''
def turn_right():
    '''Turn the robot right 90 degrees.
    Returns:
        A string message of the environment feedback.
    '''
def raise_camera():
    '''Raise the camera to see objects that are higher.
    Returns:
        A string message of the environment feedback.
    '''
def lower_camera():
    '''Lower the camera to see objects that are lower.
    Returns:
        A string message of the environment feedback.
    '''
def put_inside(obj1, obj2):
    '''Put obj1 within your hand inside obj2. If obj2 is openable, make sure it is open before putting obj1 inside.
    Args:
        :param obj1: the digital identifier of the object to put inside.
        :param obj2: the digital identifier of the object to put inside of.
    Returns:
        A string message of the environment feedback.
    '''
def put_on_top(obj1, obj2):
    '''Put obj1 within your hand to the top of obj2.
    Args:
        :param obj1: the digital identifier of the object to put on top.
        :param obj2: the digital identifier of the object to put on top of.
    Returns:
        A string message of the environment feedback.
    '''
def put_under(obj1, obj2):
    '''Put obj1 within your hand to the bottom of obj2.
    Args:
        :param obj1: the digital identifier of the object in your hand.
        :param obj2: the digital identifier of the object to put obj1 under.
    Returns:
        A string message of the environment feedback.
    '''
def put_next_to(obj1, obj2):
    '''Put obj1 within your hand next to obj2.
    Args:
        :param obj1: the digital identifier of the object in your hand.
        :param obj2: the digital identifier of the object to put obj1 next to.
    Returns:
        A string message of the environment feedback.
    '''
def get_fridge_view():
    '''Get the image captured by a camera in the fridge. This function is only valid when you are near a fridge and the fridge is open.
    Returns:
        A string message of the environment feedback.
    '''
def cook(obj):
    '''Cook the object.
    Args:
        :param obj: the digital identifier of the object to cook.
    Returns:
        A string message of the environment feedback.
    '''
def burn(obj):
    '''Burn the object.
    Args:
        :param obj: the digital identifier of the object to burn.
    Returns:
        A string message of the environment feedback.
    '''
def freeze(obj):
    '''Freeze the object.
    Args:
        :param obj: the digital identifier of the object to freeze.
    Returns:
        A string message of the environment feedback.
    '''
def heat(obj):
    '''Heat the object.
    Args:
        :param obj: the digital identifier of the object to heat.
    Returns:
        A string message of the environment feedback.
    '''
def open(obj):
    '''Open the object.
    Args:
        :param obj: the digital identifier of the object to open.
    Returns:
        A string message of the environment feedback.
    '''
def close(obj):
    '''Close the object.
    Args:
        :param obj: the digital identifier of the object to close.
    Returns:
        A string message of the environment feedback.
    '''
def toggle_on(obj):
    '''Toggle on the object.
    Args:
        :param obj: the digital identifier of the object to toggle on.
    Returns:
        A string message of the environment feedback.
    '''
def toggle_off(obj):
    '''Toggle off the object.
    Args:
        :param obj: the digital identifier of the object to toggle off.
    Returns:
        A string message of the environment feedback.
    '''
def done():
    '''Call this function if you think the task is completed. Note that you have no chance to take any actions after calling this function.
    Returns:
        None. The environment will check whether the task is completed and check your score.
    '''
```
## Reminder
1. You can only hold one object at a time.
2. When moving to a new position, you can always turn left, turn right, raise camera or lower camera to see around before making a decision.
3. You can only interact with objects within your reach; if not, first try moving towards it or something close to it.
4. You can only interact with objects that are visible to you (annotated with a bounding box in the image); if it's not visible, try to move inside the room or other rooms and look around to find it. You can open refrigerators or other enclosures to see inside them.
5. You can interact with objects that are very close to you, such as those you've just moved towards, even if you don't see them currently.
6. When you are out of the room and see nothing useful, try moving to a room.
7. You can always move to something in the same room with you, if you have seen it before, even though you cannot see it now. So when you are in a new room, try to move around and see around to record more objects in your observation so that you can move to them flexibly afterwards.
8. Don't repeat the failed action in the next round. Try to understand what went wrong and make a different decision.
9. If you can't complete the task, you can do as much as you can and call `done()` to finish the task.

# Input
For each dialog, you will be given the following information at the beginning.
1. Task Goal: The task is finished only when these conditions are met.
2. Reachable Rooms: Rooms you can move to. Please refer to them with their names provided here.
For each turn, you will be given the following information.
1. Action Feedback: Environment feedback of the last action.
2. At Hand Object: The object you are currently holding.
3. Current Room: The room you are currently in.
4. Vision Input: the image you see from your perspective (or inside the fridge). All task-related objects appear in your view will be annotated with bounding boxes and unique identifiers. Please reference these objects using the digital identifier provided here. Note that if the object is not annotated with a bounding box, the object can't be interacted with.

# Output
Now, given these information, you need to think and call the action needed to proceed with the task. Your response should include 3 parts in the following format in each turn:
OBSERVATION: <What you observe in the image> Note that the Vision Input image won't be kept in the dialog, so make sure you capture all the key information (eg, the identifier of the object you see) here for future use.
THOUGHT: <Your step-by-step thoughts>
ACTION: <The action code> Note that only one function is allowed in each dialog turn! Only one line of code is allowed in each dialog turn! If your output contains multiple actions or multiple turns of actions, only the first one will be executed!
\end{lstlisting}

Here is a concrete example of the task input shown in Fig.~\ref{fig:omnigibson_example}, where the image is enclosed within "\texttt{\{\{\}\}}":
\lstset{
    backgroundcolor=\color[RGB]{245,245,244},
    breaklines=true,
    basicstyle=\ttfamily\small
}\begin{lstlisting}
Your task is: There are 4 beer bottles on a countertop in the kitchen. Please put all of them into the fridge.
The reachable rooms during the task are: corridor_0, dining_room_0, kitchen_0, living_room_0, pantry_room_0, storage_room_0.
Action Feedback: None actions before.
At Hand Object: None.
Current Room: kitchen_0.
Vision Input: {{Image}}
\end{lstlisting}

\section{\sm-Minecraft}
\label{appendix:minecraft}
In this section, we provide additional details about \sm-Minecraft that are not covered in the main paper due to space limitations.

The game Minecraft has become a popular open-world environment for developing generalist embodied agents~\cite{fan2022minedojo,lifshitz2024steve} due to its diverse tasks (e.g., survival, harvest, craft, combat, and creative tasks), varied environments, and interactive mobs, necessitating generalized agent abilities. Recent pioneering works~\cite{zhu2023ghost,wang2023voyager,wang2023jarvis} have integrated LLMs into embodied agents to tackle Minecraft tasks. However, these efforts did not focus on a standardized pipeline for evaluating LMMs' planning abilities. So we adapt the JARVIS-1~\cite{wang2023jarvis} pipeline to assess LMMs' high-level action planning abilities in item-obtaining tasks.

\subsection{Detailed Description}
In \sm-Minecraft, we adapt the action space of JARVIS-1~\cite{wang2023jarvis} to develop a pipeline for LMM, enabling it to function as a high-level embodied planner. Additionally, we also use item-obtaining tasks to benchmark LMMs' high-level embodied planning abilities. These tasks are comprehensive, requiring task analysis and decomposition, as well as ingredient collection. Each aspect respectively challenges an LMM agent's planning skills and its ability to interact with the environment.

\vpara{Test Set.}
We manually annotate 116 different tasks, each with a specific target item and a corresponding initial configuration to ensure the task is solvable. For example, Fig.~\ref{fig:minecraft_example} illustrates the \sm-Minecraft task of obtaining a cake, where we have set up the initial configuration of necessary surrounding resources and inventory items. These 116 test tasks span the Minecraft tech tree, covering items across 6 material levels (wood, stone, iron, gold, diamond and netherite) and involving a diverse range of raw ingredients from various resources: 11 types of plants, 4 types of animals, and 6 types of hostile mobs. This diversity greatly challenges the agent's ability to interact with the environment.

\vpara{Training Set.}
Training trajectories are collected using bootstrapping from two sources: pure \texttt{gpt-4-turbo} bootstrapping on newly designed tasks, and \texttt{gpt-4o} bootstrapping with JARVIS-1 memory on tasks from JARVIS-1. For the first type, we design 40 new tasks instantiated in different world seeds or spawn points, resulting in 512 task instances, and \texttt{gpt-4-turbo} bootstraps 176 successful trajectories. For the second type, we use 132 tasks from JARVIS-1, set up in 596 task instances, and run with memory using \texttt{gpt-4o}, resulting in 206 successful trajectories. In total, we gain 382 successful trajectories.

\vpara{Metrics.}
We adopt success rate as the evaluation metric in \sm-Minecraft. For a specific item-obtaining task, if the agent can obtain the specific item within the limitation of 100 rounds, the task is regarded as successfully completed.

\subsection{Actions}
In \sm-Minecraft, we provide 6 types of actions for the LMM agent. 4 actions, marked with an asterisk (*), are adapted from the JARVIS-1 pipeline~\cite{wang2023jarvis}, while the remaining 2 are newly implemented by us to enhance the LMM agent's capability to solve a wider range of tasks.

\begin{itemize}[leftmargin=1.5em,itemsep=0pt,parsep=0.2em,topsep=0.0em,partopsep=0.0em]
\item \textbf{craft}*: Utilize the inventory or crafting table to craft a specific item.
\item \textbf{smelt}*: Utilize a furnace to smelt a specific item.
\item \textbf{equip}*: Equip a specific item in the agent's hand.
\item \textbf{teleport\_to\_spawn}: Teleport the agent back to the spawn point. As we will prepare necessary ingredients around the agent's spawn point, this action enables the agent to conveniently collect these ingredients. This function is also helpful if the agent stuck somewhere (e.g., underground).
\item \textbf{look\_up}: Look up the crafting/smelting information about a specific item. This reference guides the agent to make a plan on how to accomplish the task.
\item \textbf{execute}*: Use natural language prompt to instruct a low-level minecraft planner, Steve-1~\cite{lifshitz2024steve}. With proper prompting, it can solve most basic tasks, like mining common blocks, collecting plants, interacting with animals and hostile mobs, and navigating between different biomes.
\end{itemize}

\subsection{Prompt Example}

\begin{figure}[t]
    \centering
    \includegraphics[width=0.9\textwidth]{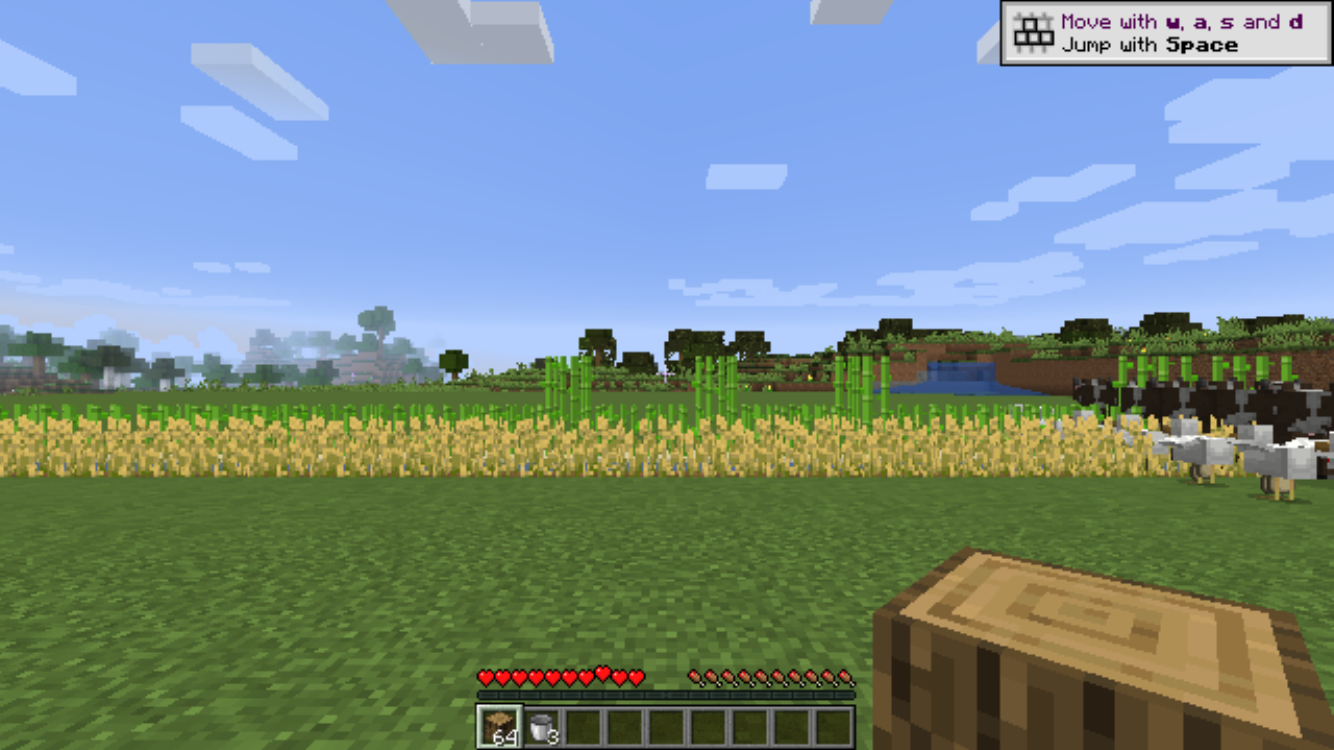}
    \caption{This is an example of \sm-Minecraft task. This task asks the agent to obtain a cake in the inventory. Initially, we provide 3 buckets and 64 logs in the inventory. Additionally, we grow mature wheat and sugar cane in front of the agent and spawn a few chickens and cows around it, ensuring that the agent can conveniently find the necessary ingredients.}
    \label{fig:minecraft_example}
\end{figure}

The system message that describes the detailed task information to the agent is shown as follows:
\lstset{
    backgroundcolor=\color[RGB]{245,245,244},
    breaklines=true,
    basicstyle=\ttfamily\small
}\begin{lstlisting}
# Setup
You are a skilled Minecraft player. You are born in the survival mode and asked to obtain a specific item.
You can interact with the game environment by outputing actions using python-style pseudo code. For each turn, please call exactly one predefined function.

# Valid Actions
## Predefined Function List:
```
def craft(item: str, num: int = 1):
    '''Craft specified number of items. Please ensure that you get enough ingredients and a craft_table in your inventory.
    Args:
        obj: the name of the item to craft.
        num: the number of items to craft. Default is 1.
    Returns:
        A string message about whether the crafting is successful.
    Examples:
        >>> craft("wooden_pickaxe")
        Successfully crafted 1 wooden_pickaxe.  
        >>> craft("bookshelf", 2)
        Not enough materials for 2 bookshelf.   # You don't have 12 planks and 6 books in your inventory.
    '''

def smelt(item: str, num: int = 1):
    '''Smelt specified number of items. Please ensure that you get enough fuels, ingredients, a furnace and a **wooden_pickaxe** in your inventory.
    Args:
        obj: the name of the item to smelt.
        num: the number of items to smelt. Default is 1.
    Returns:
        A string message about whether the smelting is successful.
    Examples:
        >>> smelt("iron_ingot", 2)
        Successfully smelted 2 iron_ingot.
        >>> smelt("glass")
        Not enough fuels.  # You don't have enough coals, logs or planks as fuel.
    '''

def equip(item: str):
    '''Select an item from your inventory to your hand. Note that if you want to use some item, you must equip it first!
    Args:
        item: the name of the item to equip.
    Returns:
        A string message about whether the equipping is successful.
    Examples:
        >>> equip("diamond_sword")
        Successfully equipped diamond_sword.
        >>> equip("diamond_axe")
        Can not find diamond_axe in inventory.  # You must have the item in your inventory before equipping it.
    '''

def teleport_to_spawn():
    '''teleport yourself to the spawn position.
    Args:
        None.
    Returns:
        A string message about whether the teleportation is successful.
    Examples:
        >>> teleport_to_spawn()
        Successfully teleported.

def look_up(item: str):
    '''Look up the information about crafting the item.
    Args:
        item: the name of the item/tag to look up.
    Returns:
        A string message about the information of the item. Note that if the argument is a tag, information about all possible items will be returned.
    Examples:
        >>> look_up("iron_pickaxe")
        iron_pickaxe: Crafting iron_pickaxe needs 2 stick, 3 iron_ingot. Every time you craft iron_pickaxe with the ingredients above, you will get 1 iron_pickaxe.
        >>> look_up("stone_tool_materials")
        stone_tool_materials is a tag. Following items belong to this tag: cobblestone, blackstone.
        cobblestone: It is a raw item you can mine from the environment.
        blackstone: It is a raw item you can mine from the environment.
    '''

def execute(prompt: str, goal_item: Optional[str] = None, num: Optional[int] = None)
    '''Instruct a lower-level executor model to perform some simple tasks, like mine something, collect something, move to some place.
    Args:
        prompt: the prompt to instruct the lower-level executor model. It should be a simple **verb-object phrase**.
        goal_item (optional): the name of the item to obtain during the execution. If the item is obtained, the executor model will stop.
        num (optional): the number of items to obtain.
    Returns:
        A string message about the execution.
    Negative Examples: # examples that may cause failure
        Your Inventory: Now your inventory has 1 stone_pickaxe, 2 stick.
        Equipped Item: Now you hold the stone_pickaxe in your hand.
        >>> execute("break iron_ore blocks", "iron_ore", 64)
        The executor has reached the maximum number of steps for this turn without completing your subgoal. # Each turn is limited in time, 64 iron_ore is too much for one turn.

        Your Inventory: Now your inventory has 1 wooden_axe, 12 logs, 4 stick, 1 seed, 1 iron_pickaxe.
        Equipped Item: Now you hold the wooden_axe in your hand.
        >>> execute("find and mine diamond", "diamond_ore", 1)
        The executor has reached the maximum number of steps for this turn without completing your subgoal. # You are not holding the right tool for mining diamonds. You should equip the iron_pickaxe first.

        Your Inventory: Now your inventory has 64 dirt.
        Equipped Item: Now you hold nothing in your hand.
        >>> execute("climb on a tree")
        The executor has attempted to execute the action according to your prompt. You should check whether your intention has been fulfilled. # The executor can't plan for complex tasks; you have to break down complex tasks into simple ones. For example, break down the task of `climb on a tree` into `find a tree`, `use dirt blocks to elevate`, and `jump on the tree`.

        Your Inventory: Now your inventory has nothing.
        Equipped Item: Now you hold nothing in your hand.
        >>> execute("dig a hole and jump in")
        Error: No complex sentences allowed. Keep the prompt a simple **verb-object phrases**. # Your prompt contains multiple tasks that may be confusing to the executor.

        Your Inventory: Now your inventory has 4 logs.
        Equipped Item: Now you hold nothing in your hand.
        >>> execute("craft a wooden_axe", "wooden_axe", 1)
        Error: You cannot use `execute` to craft items. Use `craft` instead. # The executor cannot craft or smelt items, call `craft` for `smelt` function instead.

        Your Inventory: Now your inventory has 4 logs, 1 crafting_table.
        Equipped Item: Now you hold nothing in your hand.
        >>> execute("place crafting_table")
        Error: You cannot use `execute` to craft items or place the crafting_table. Directly use `craft` instead. No need to place the crafting_table. # The `craft` function will automatically place the crafting_table during crafting.

        Your Inventory: Now your inventory has nothing.
        Equipped Item: Now you hold nothing in your hand.
        >>> execute("hold down left button to punch the tree to collect wood", "logs", 1)
        The executor has reached the maximum number of steps for this turn without completing your subgoal. # The description of the task is too complex, it should be a **verb-object phrase**.

    Positive Examples: # good examples for reference
        Your Inventory: Now your inventory has stone_pickaxe, stick.
        Equipped Item: Now you hold the stone_pickaxe in your hand.
        >>> execute("break iron_ore blocks", "iron_ore", 2)
        Your subgoal has been successfully completed by the executor. # You have seen the iron_ore and you are using the correct tool. Note that if you haven't seen the iron_ore, you'd better use `break stone, obtain iron ore` as your prompt.

        Your Inventory: Now your inventory has nothing.
        Equipped Item: Now you hold nothing in your hand.
        >>> execute("collect wood", "logs", 1)
        Your subgoal has been successfully completed by the executor. # The executor can only understand the instructions of simple **verb-object phrases**.

        Your Inventory: Now your inventory has nothing.
        Equipped Item: Now you hold nothing in your hand.
        >>> execute("dig a hole", "dirt", 4)
        Your subgoal has been successfully completed by the executor. # Your instructions are simple and easy to understand.

        Your Inventory: Now your inventory has 1 wooden_axe, 2 stick.
        Equipped Item: Now you hold the wooden_axe in your hand.
        >>> execute("find a river")
        The executor has attempted to execute the action according to your prompt. You should check whether your intention has been fulfilled. # The executor has the ability to find the environment you are looking for, despite the possibility of failure.

    Prompt Examples: # some simple prompts for reference
    "chop down the tree", "break leaves", "collect seeds", "break a flower", "dig down", "break stone, obtain iron ore", "break gold_ore blocks", "mine diamond ore", "kill sheep", "milk cow", "combat spider", "find a river", "break stones", "break sand blocks", "move out of the cave".
    '''
```
## Reminder
1. You can only call one function in each turn.
2. If you have no idea on how to solve the task or are unfamiliar with some items, please call the `look_up` function to check the item.
3. For some items that you can not mine or obtain with your bare hand, try to equip a pickaxe (wooden_pickaxe, stone_pickaxe, ...) before mining it.
4. Some necessary resources (e.g., mobs, plants) might be prepared for you near the spawn point. If you're struggling to find certain ingredients or find yourself stuck somewhere, you can use the `teleport_to_spawn` function to return there.
5. When calling the executor, keep the positive examples and negative examples in mind! If the executor cannot complete your subgoal, check whether you have the right item in your hand, and try to break your prompt into smaller steps and adjust your subgoal, modify the prompt, or carefully repeat the prompt.
6. Do not repeat the failed action in the next round. Try to understand what went wrong and make a different decision.

# Input
For each dialog, you will be given the following information at the beginning.
- Task Goal: The item you should obtain in your inventory.
For each turn, you will be given the following information.
1. Feedback on the Action: The feedback on the action you output in the last turn.
2. Your Inventory: The items in your inventory.
3. Equipped Item: The item you are currently holding in your hand.
4. Location and Orientation: including X, Y, Z, Pitch and Yaw. X and Z are horizontal coordinates; Y is the height. Pitch measures the tilt of the player's view: 0, positive values and negative values mean the player is looking horizontally, downward, and upward, respectively. Yaw measures the rotation around the player's vertical axis: 0 or 360 degrees north, 90 degrees east, 180 degrees south, and 270 degrees west.
5. Vision Input: What you see from your perspective.

# Output
Now, given these information, you need to think and call the action needed to proceed with the task. Your response should include 3 parts in the following format in each turn:
OBSERVATION: <What you observe in the image> Note that the Vision Input image won't be kept in the dialog, so make sure you capture all the key information (eg, the biome or items you see) here for future use.
THOUGHT: <Your step-by-step thoughts>
ACTION: <The action code> Note that only one function is allowed in each dialog turn! Only one line of code is allowed in each dialog turn! If your output contains multiple functions or multiple turns of functions, only the first one will be executed!
\end{lstlisting}

Here is a concrete example of the task input shown in Fig.~\ref{fig:minecraft_example}, where the image is enclosed within "\texttt{\{\{\}\}}":
\lstset{
    backgroundcolor=\color[RGB]{245,245,244},
    breaklines=true,
    basicstyle=\ttfamily\small
}\begin{lstlisting}
Your task is to get a cake in your inventory.
Feedback on the Action: No action before.
Your Inventory: Now your inventory has 64 oak_log, 3 bucket.
Equipped Item: Now you hold the oak_log in your hand.
Location and Orientation: Now you locate in X: 431.50, Y: 65.00, Z: -158.50, Pitch: 0.00, Yaw: 0.00.
Vision Input: {{Image}}
\end{lstlisting}
\section{\sm-Mobile}
\label{appendix:mobile}
In this section, we provide additional details regarding \sm-Mobile that are not covered in the main text due to space limitations.

\subsection{Detailed Description}
To introduce the Android Eval benchmark, we developed a framework including an operational environment and a benchmark tailored for agents interacting with Android.

Android Eval benchmark comprises 119 tasks across 8 different apps, offering evaluation suites considering the device's and screen's state. It implements evaluation frameworks for both the ReAct~\cite{yao2022react} and SeeAct~\cite{zheng2024gpt4vision} methods. For reproducibility, the Android virtual device provides standard evaluation virtual machines preloaded with various apps' operation histories and offline data, ensuring that network or temporal factors do not affect evaluations. To simulate real-world tasks, we offer Android virtual machine images with randomized operations, ensuring evaluations do not have to start from an initial usage state and enabling more complex task completion recognition based on the machine and current page state.

\subsection{Actions}
In \sm-Mobile, agents are required to accomplish diverse user tasks through predefined actions. 

\begin{itemize}[leftmargin=1.5em,itemsep=0pt,parsep=0.2em,topsep=0.0em,partopsep=0.0em]
    \item \textbf{tap}: Tap element with specific id.
    \item \textbf{type}: Type the message into the input box and press enter if needed.
    \item \textbf{long press}: Tap element with specific id for a long duration.
    \item \textbf{swipe}: Swipe with distance and direction.
    \item \textbf{finish}: Finish the task with optional message.
    \item \textbf{press back}: Press back button.
    \item \textbf{press home}: Press home button.
\end{itemize}

\subsection{Metrics}
The metric we designed is directly oriented towards task completion. We can directly assess the task's success rate by checking whether the operation sequence includes necessary screens or device states that indicate task completion. For example, in setting an alarm time, we sequentially check if the task sequence includes the correctly set alarm time and if the alarm is turned on. Specifically, the metrics we designed are as follows:
\begin{itemize}[leftmargin=1.5em,itemsep=0pt,parsep=0.2em,topsep=0.0em,partopsep=0.0em]
    \item \textbf{Success Rate}: We measure the success rate by device state and screen state for the operation task.We measure the success rate for the query task by comparing the model answer with the ground truth.
\end{itemize}

\subsection{Prompt Example}
Here is the system prompt we use.
\lstset{
    backgroundcolor=\color[RGB]{245,245,244},
    breaklines=true,
    basicstyle=\ttfamily\small,
    escapeinside=||
}\begin{lstlisting}
You are an agent that is trained to complete certain tasks on a smartphone. You will be 
given a screenshot of a smartphone app. The interactive UI elements on the screenshot are labeled with numeric tags 
starting from 1. 

You can call the following functions to interact with those labeled elements to control the smartphone:

1.tap(index: int)

Taps the UI element labeled with the given number.
Example: tap(5)

2.text(input_str: str)

Inserts the given text into an input field. 
Example: text("Hello, world!")
Since we use ADB keyboard, if ADB keyboard ON is displayed on the bottom of the screen, you can use this function.
If you think that the keyboard is displayed after your previous operation, you can try to use this function to input text.

3.long_press(index: int)

Long presses the UI element labeled with the given number.
Example: long_press(5)

4. swipe(index: int, direction: str, dist: str)

Swipes the UI element in the specified direction and distance. "direction" is a string that 
represents one of the four directions: up, down, left, right. "dist" determines the distance of the swipe and can be one
of the three options: short, medium, long.
Example: swipe(21, "up", "medium")

5. back()

Simulates a back button press on the smartphone.

6. home()

Simulates a home button press on the smartphone.

7. wait(interval: int)

Pauses the execution for the given number of seconds. Default is 5 second.

8. finish(message: str)

Ends the task and provides the final output. You can return the final output of the task as a string.
Example: finish("Task completed")

Now, given the following labeled screenshot, you need to think and call the function needed to proceed with the task. 
Your output should include only action part in the given format:

Action: <The function call with the correct parameters to proceed with the task. If you believe the task is completed or 
there is nothing to be done, you should use finish function. You cannot output anything else except a function call
in this field.>

Whenever you think the task is finished, you should use finish function to avoid extra operations.

If you found yourself in a loop or the task is not proceeding as expected, you might consider changing your operation and try other methods.
If you operate same action 5 times, the program will automatically stop.
If tap operation is not working, you can try long press operation.

You can only take one action at a time, so please directly call the function.
\end{lstlisting}
\section{WebArena-Lite}
\label{appendix:webarena}
In this section, we provide additional details regarding WebArena-Lite that are not covered in the main text due to space limitations.

\subsection{Detailed Description}
WebArena~\cite{zhou2023webarena} is designed to evaluate the ability of agents to perform complex user tasks described in high-level natural language in a realistic, interactive web environment. To achieve this goal, WebArena presented a highly simulated and interactive web environment, which consists of five common websites, including Gitlab, map, forum, online shopping, and content management platform. It is also equipped with external tools such as sketch pad and calculator, which enhance the ability of the agents to perform user tasks.
In contrast to other benchmarks where the agents are constrained to act as website users, WebArena proposed innovative ways to simulate different user roles. For instance, they constructed a content management platform (CMS) and granted the agent full administrative privileges.  This assesses the agent's capacity to assume various roles in complex scenarios.

\begin{itemize}[leftmargin=1.5em,itemsep=0pt,parsep=0.2em,topsep=0.0em,partopsep=0.0em]
    \item \textbf{Task Description:} As web GUI agents, LMMs are asked to accomplish user instructions on certain websites. For example, on \texttt{OneStopShop} website, an instruction would be ``\texttt{What do customers say about brush from sephora}'', and LMM agents should search for the product, enter the review section, and summarize the customer reviews (or turn out finding no review). To enable the action of LMM agents with visual input, we implement HTML SoM~\cite{koh2024visualwebarena} to annotate operable HTML elements with ids on the screenshot, we also provide a list of textual information for all clickable elements. LMM agents generate actions and the id of elements being operated by \texttt{playwright}.
    \item \textbf{Test Set:} We build WebArena-Lite, a subset of 165 representative tasks by selection, refinement, and adaptation to multimodality evaluation (i.e., screenshot). Our refinement focuses on resolving implausible judge conditions, where 30 tasks are being manually fixed (Cf. Appendix~\ref{appendix:webarena-fix}). The implausibility may involve wrong answers, too-strict criteria (e.g., \texttt{exact\_match}), impossible tasks due to environment bugs, etc. Additionally, we remove cross-website tasks for simplicity of testing.
    \item \textbf{Training Set:} Creating environment-dependent task instructions and trajectories for training on web could be challenging. In \sm, for each website we first summarize the basic functions and valid items for synthetic queries to condition on. Based on summarized functions, we come up with an array of task prototypes (with item placeholders) and manually write \texttt{playwright} scripts as rule-based solvers for each task prototype. We fill task prototypes with both valid and invalid items to yield detailed instructions (later being rephrased by LLMs for expression diversity), and run corresponding solvers on the website to collect groundtruth trajectories with screenshots and operations. 5 authors create around 40 task prototypes with corresponding solvers, and generating 1,186 valid training samples (i.e., instruction, trajectory, and reward function) for WebArena-Lite.
\end{itemize}

\subsection{Actions}
In WebArena-Lite, agents are required to accomplish diverse user tasks through a series of predefined actions. However, real-world webpages are often complex, and thus, we provide these actions in order to ensure simplicity and practicality.

\begin{figure}[t]
    \centering
    \includegraphics[width=0.9\textwidth]{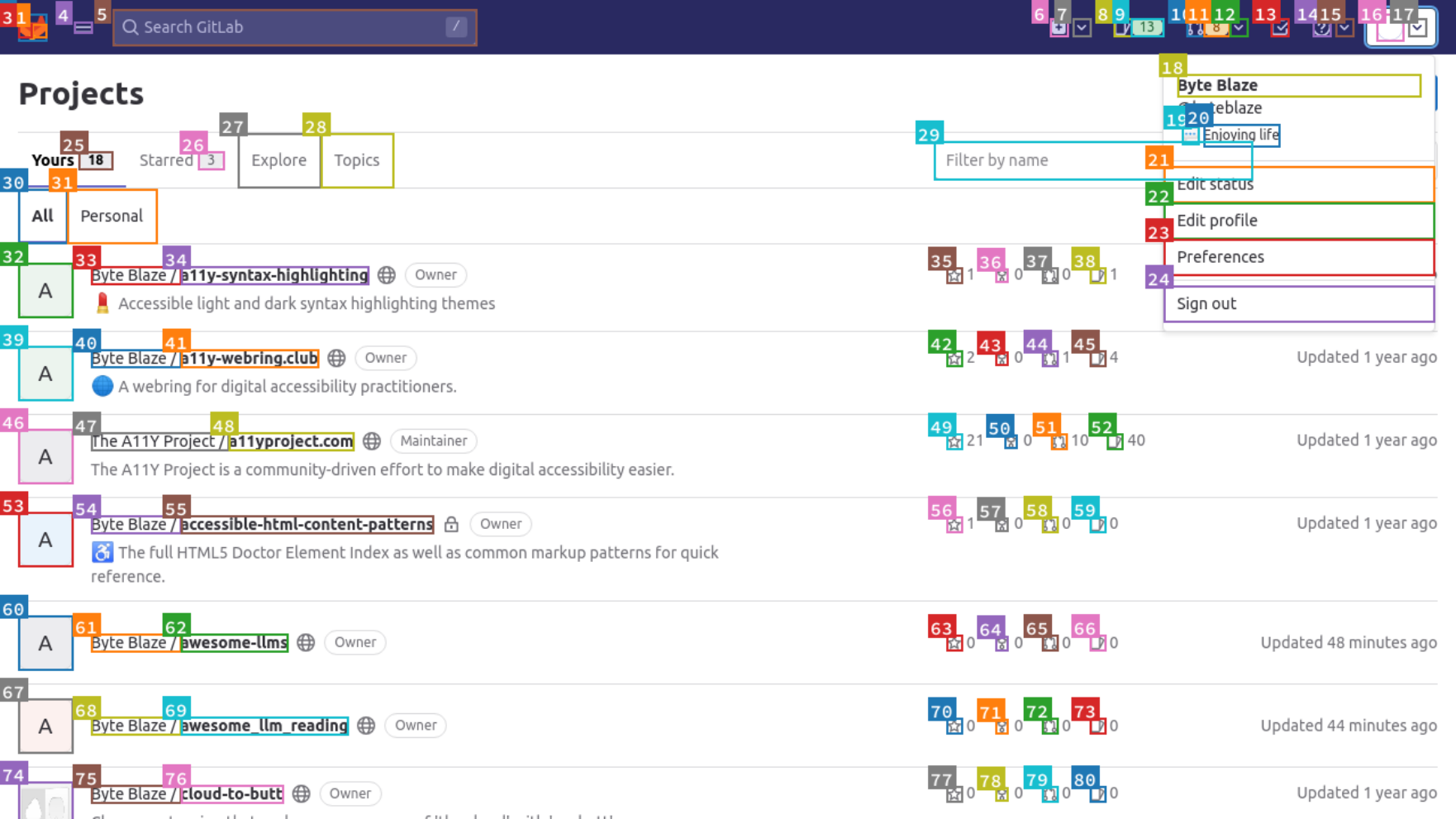}
    \caption{This is an example of WebArena-Lite task where we use the SoM approach to highlight actionable elements. This task requires the agent to modify the user's status information. To accomplish this, the agent initially clicks on the user's avatar, which directs them to the status shown in the figure. At this point, the agent should select the option labeled "\texttt{(21) Edit Status}" in order to access the modification page and complete the task.}
    \label{fig:webarena_example}
\end{figure}

\begin{itemize}[leftmargin=1.5em,itemsep=0pt,parsep=0.2em,topsep=0.0em,partopsep=0.0em]
    \item \textbf{click}: Click element with specific id.
    \item \textbf{hover}: Hover element with specific id.
    \item \textbf{type}: Type the message into the input box with a specific id and press enter if needed.
    \item \textbf{press}: Emulates a keyboard key combination.
    \item \textbf{scroll}: Scrolls the page up or down.
    \item \textbf{new\_tab}: Opens a new tab in the current browser.
    \item \textbf{tab\_focus}: Switches to the tab with specific index.
    \item \textbf{close\_tab}: Closes the current tab.
    \item \textbf{goto}: Go to specific URL.
    \item \textbf{go\_back}: Go back to the previous page.
    \item \textbf{go\_forward}: Go to the next page if it exists.
    \item \textbf{stop}: Terminates the operation, returns the response, and exits. 
\end{itemize}

\subsection{Metrics}
In real-world web browsing scenarios, there can be multiple ways for an agent to accomplish a task. Consequently, WebArena-Lite only considers whether the task has been completed or not, without considering the execution trajectory of the agent, therefore, the metric used in WebArena-Lite is \textbf{Success Rate (SR)}. We maintain the evaluation method described by WebArena~\cite{zhou2023webarena}, which can be categorized into three categories based on task type.

\begin{itemize}[leftmargin=1.5em,itemsep=0pt,parsep=0.2em,topsep=0.0em,partopsep=0.0em]
    \item \textbf{Question Answering}: Agent needs to give an answer and the score depends on the string-matching result.
    \item \textbf{Webpage Navigation}: Agent must navigate to a specific web page. The completion of the task is dependent on the URL of the page on which the agent terminated.
    \item \textbf{Content modification}: Agent needs to interact with the environment to modify the configuration of the webpage, and the evaluation will extract the content of the page and match it to check whether the content meets the expectations.
\end{itemize}

In light of the aforementioned considerations, string-matching patterns can be classified into three distinct categories:

\begin{itemize}[leftmargin=1.5em,itemsep=0pt,parsep=0.2em,topsep=0.0em,partopsep=0.0em]
    \item \textbf{exact\_match}: The response of the agent is scored when it exactly matches the token sequence corresponding to the answer.
    \item \textbf{must\_include}: Answers that contain a specific token sequence are considered a match.
    \item \textbf{fuzzy\_match}: Utilizes LLMs such as GPT-4 to assist in determining whether an answer is correct.
\end{itemize}

The selection of appropriate evaluation metrics for distinct types of tasks enables the construction of a comprehensive and relatively accurate test set.

\subsection{Task Amendment}\label{appendix:webarena-fix}
Some tasks in WebArena have typos, incorrect answers, and inaccurate scoring criteria. Therefore, we selected 165 tasks from WebArena with different templates and then corrected 39 of them, as shown in Table~\ref{tab:webarena_lite}. Considering that the model uses natural language to answer the questions, we change the tasks that require exact match to must include or fuzzy match, and also correct the answers.

{
\small
\renewcommand\arraystretch{1.25}
\renewcommand\tabcolsep{3pt}
\begin{longtable}{llp{7cm}<{\raggedright}p{2.15cm}<{\raggedright}p{2.15cm}<{\raggedright}}
\caption{Task instructions fixed in WebArena-Lite} \\

\toprule
ID & Website & Task & Before & After \\
\hline
\endhead

\endfoot

\bottomrule
\endlastfoot

7 & Map & Tell me the full address of all international airports that are within a driving distance of 50 km to Carnegie Mellon University. & exact\_match & fuzzy\_match \\ \hline
33 & Map & I will arrive Pittsburgh Airport soon. Provide the name of a Hilton hotel in the vicinity, if available. Then, tell me the the shortest walking distance to a supermarket from the hotel. & must\_include & fuzzy\_match \\ \hline
37 & Map & Check if the police station in pittsburgh can be reached in one hour by car from gates building at CMU. & must\_include & fuzzy\_match \\ \hline
43 & CMS & List the top 3 search terms in my store. & hollister, Joust Bag, Antonia Racer Tank & hollister, Joust Bag, nike \\ \hline
65 & CMS & Which customer has completed the fifth most number of orders in the entire history? & Jane Doe & Matt Baker \\ \hline
71 & Map & What is the zip code of Chatham University? & exact\_match & must\_include \\ \hline
82 & Map & What is the duration required to first walk from Massachusetts Institute of Technology to Harvard University, and then drive to Boston Logan International Airport? & 63 min & 64 min \\ \hline
97 & Map & Tell me the distance to drive from Carnegie Mellon University to the top computer science school in massachusetts. & must\_include & fuzzy\_match \\ \hline
98 & Map & Where is the nearest tea cafe to University of Pittsburgh, and what is the walking distance to it? & must\_include & fuzzy\_match \\ \hline
103 & Gitlab & Display the list of issues in the kkroening/ffmpeg-python repository that have labels related to questions. &  & URL: sort by created\_date, state is opened \\ \hline
109 & CMS & Presents the monthly count of successful orders \{\{period\}\} in MM:COUNT format. & January: 11 orders, February: 16 orders & 01:11, 02:16 \\ \hline
127 & CMS & What brands appear most frequently among the top search terms? & Hollister, Joust, Antonia & Hollister \\ \hline
135 & Gitlab & How many commits did Eric and Kilian make to a11yproject on 1/3/2023? & 1 & 0 \\ \hline
167 & OSS & What are the main criticisms of this product? Please extract the relevant sentences. & must\_include & fuzzy\_match \\ \hline
215 & CMS & What are the key aspects that the customers don't like about Circe ice fleece. & fuzzy\_match (“Material quality, …”) & exact\_match (“N/A”) \\ \hline
225 & OSS & What do customers say about brush from sephora. & N/A & No reviews available \\ \hline
235 & OSS & Get the order number of my most recent under delivery order. & fuzzy\_match & must\_include \\ \hline
236 & Map & Where is the nearest pharmacy from Carnegie Mellon I can walk within 20mins. & must\_include & fuzzy\_match \\ \hline
240 & OSS & I am doing a market survey for one stop market, show me the most expensive product from competative swimwear category. &  & URL \\ \hline
254 & Map & What is the phone number of Western Pennsylvania Hospital. & exact\_match & fuzzy\_match \\ \hline
259 & Gitlab & Get me my RSS feed token. & exact\_match & must\_include \\ \hline
287 & Map & How much time does it take from Pittsburgh to Philadelphia by car? & 5h 47m & 8h 33m \\ \hline
288 & CMS & Tell me the name of the customer who has the most cancellations in the history. & exact\_match & must\_include \\ \hline
296 & Gitlab & Show me the command to clone the best GAN python implementation with SSH. & metis.lti. cs.cmu.edu & localhost \\ \hline
311 & Gitlab & Tell me who has made the most contributions, in terms of number of commits, to the Pytorch GAN project. & exact\_match & must\_include \\ \hline
348 & CMS & How many reviews our shop received in May 2023? & must\_include & fuzzy\_match \\ \hline
349 & Gitlab & Who else have access to my repo gimmiethat.space, show me their usernames. & exact\_match & must\_include \\ \hline
367 & Map & Measure distance between Carnegie Mellon University and CVS (closet one) by walking. & exact\_match & fuzzy\_match \\ \hline
376 & OSS & Summarize customer reviews for Amazon Echo Dot 3rd generation. & N/A & fuzzy\_match (“Serena: it…”) \\ \hline
392 & Gitlab & Post \textbackslash{}"Good idea\textbackslash{}" for the merge request related to color ulitity in a11yproject.com project”. &  & URL \\ \hline
528 & OSS & Draft a refund message via their \textbackslash{}"contact us\textbackslash{}" form for the phone screen protector I bought March 2023. It broke after three days of use. The shop requires the order id, the reason and the amount to refund in the message. Don't submit yet. & must\_include & fuzzy\_match \\ \hline
538 & CMS & Modify the address of order \#299 to 456 Oak Avenue, New York, NY, 10001. & …456 Oak Avenue, Apartment 5B, New York… & …456 Oak Avenue, New York… \\ \hline
641 & Reddit & Post a notice on a virtual meetup for Harry Potter enthusiasts on July 8th in the books subreddit. & must\_include & fuzzy\_match \\ \hline
645 & Reddit & Post in books subreddit about what could machine learning help the correpong field. & must\_include & fuzzy\_match \\ \hline
657 & OSS & Fill the \textbackslash{}"contact us\textbackslash{}" form in the site for a refund on the speaker I bought, stating that it broke after just three days of use. Also, ensure to include the order number \#148 and the product SKU. Don't submit yet, I will check. & must\_include & fuzzy\_match \\ \hline
668 & Gitlab & Submit a merge request for a11yproject.com/redesign branch to be merged into master branch, assign Roshan Jossy as the reviewer. & Justin Armstrong & Roshan Jossy \\ \hline
693 & OSS & Draft an email to the shop owner via their contact us function for a coupon as my refund is suppoed to be replaced by a coupon. & program\_match & url\_match \\ \hline
798 & OSS & Change the delivery address for my most recent order to 77 Massachusetts Ave, Cambridge, MA. & fuzzy\_match & must\_include \\
\end{longtable}
\label{tab:webarena_lite}
}

\subsection{Prompt Example}
Here is the system prompt we use, you can find more prompt examples in VisualWebArena~\cite{koh2024visualwebarena}.
\lstset{
    backgroundcolor=\color[RGB]{245,245,244},
    breaklines=true,
    basicstyle=\ttfamily\small,
    escapeinside=||
}\begin{lstlisting}
You are an autonomous intelligent agent tasked with navigating a web browser. You will be given web-based tasks. These tasks will be accomplished through the use of specific actions you can issue.

Here's the information you'll have:
The user's objective: This is the task you're trying to complete.
The current web page's accessibility tree: This is a simplified representation of the webpage, providing key information.
The current web page's URL: This is the page you're currently navigating.
The open tabs: These are the tabs you have open.
The previous action: This is the action you just performed. It may be helpful to track your progress.

The actions you can perform fall into several categories:

Page Operation Actions:
|\texttt{\`}\texttt{\`}\texttt{\`}|click [id]|\texttt{\`}\texttt{\`}\texttt{\`}|: This action clicks on an element with a specific id on the webpage.
|\texttt{\`}\texttt{\`}\texttt{\`}|type [id] [content]|\texttt{\`}\texttt{\`}\texttt{\`}|: Use this to type the content into the field with id. By default, the "Enter" key is pressed after typing unless press_enter_after is set to 0, i.e., |\texttt{\`}\texttt{\`}\texttt{\`}|type [id] [content] [0]|\texttt{\`}\texttt{\`}\texttt{\`}|.
|\texttt{\`}\texttt{\`}\texttt{\`}|hover [id]|\texttt{\`}\texttt{\`}\texttt{\`}|: Hover over an element with id.
|\texttt{\`}\texttt{\`}\texttt{\`}|press [key_comb]|\texttt{\`}\texttt{\`}\texttt{\`}|:  Simulates the pressing of a key combination on the keyboard (e.g., Ctrl+v).
|\texttt{\`}\texttt{\`}\texttt{\`}|scroll [down]|\texttt{\`}\texttt{\`}\texttt{\`}| or |\texttt{\`}\texttt{\`}\texttt{\`}|scroll [up]|\texttt{\`}\texttt{\`}\texttt{\`}|: Scroll the page up or down.

Tab Management Actions:
|\texttt{\`}\texttt{\`}\texttt{\`}|new_tab|\texttt{\`}\texttt{\`}\texttt{\`}|: Open a new, empty browser tab.
|\texttt{\`}\texttt{\`}\texttt{\`}|tab_focus [tab_index]|\texttt{\`}\texttt{\`}\texttt{\`}|: Switch the browser's focus to a specific tab using its index.
|\texttt{\`}\texttt{\`}\texttt{\`}|close_tab|\texttt{\`}\texttt{\`}\texttt{\`}|: Close the currently active tab.

URL Navigation Actions:
|\texttt{\`}\texttt{\`}\texttt{\`}|goto [url]|\texttt{\`}\texttt{\`}\texttt{\`}|: Navigate to a specific URL.
|\texttt{\`}\texttt{\`}\texttt{\`}|go_back|\texttt{\`}\texttt{\`}\texttt{\`}|: Navigate to the previously viewed page.
|\texttt{\`}\texttt{\`}\texttt{\`}|go_forward|\texttt{\`}\texttt{\`}\texttt{\`}|: Navigate to the next page (if a previous 'go_back' action was performed).

Completion Action:
|\texttt{\`}\texttt{\`}\texttt{\`}|stop [answer]|\texttt{\`}\texttt{\`}\texttt{\`}|: Issue this action when you believe the task is complete. If the objective is to find a text-based answer, provide the answer in the bracket.

Homepage:
If you want to visit other websites, check out the homepage at http://homepage.com. It has a list of websites you can visit.
http://homepage.com/password.html lists all the account name and password for the websites. You can use them to log in to the websites.

To be successful, it is very important to follow the following rules:
1. You should only issue an action that is valid given the current observation
2. You should only issue one action at a time.
3. You should follow the examples to reason step by step and then issue the next action.
4. Generate the action in the correct format. Start with a "In summary, the next action I will perform is" phrase, followed by action inside |\texttt{\`}\texttt{\`}\texttt{\`}||\texttt{\`}\texttt{\`}\texttt{\`}|. For example, "In summary, the next action I will perform is |\texttt{\`}\texttt{\`}\texttt{\`}|click [1234]|\texttt{\`}\texttt{\`}\texttt{\`}|".
5. Issue stop action when you think you have achieved the objective. Don't generate anything after stop.
\end{lstlisting}
\section{\sm-CSS}
\label{appendix:css}
In this section, we provide additional details regarding \sm-CSS that are not covered in the main text due to space limitations.

\subsection{Detailed Description}
\label{appendix:css_description}
Existing datasets for frontend design have two major shortcomings: 
1) They focus mainly on single-round interactions, and
2) They do not provide definitive success metrics for individual tasks.
Instead, these benchmarks assess using continuous metrics like CLIP score~\cite{design2code} or qualitative analysis only~\cite{websight}.
The reason is that they expect the model to output an entire HTML file replicating the target web design, which is too challenging and unrealistic for current LMMs. 
Therefore, employing a definitive success rate as the metric is meaningless for them.
Consequently, they may fail to adequately assess LMMs' potential in serving as adaptive agents that can make new decisions based on varying observations.
Also, a binary success rate is often more decisive and crucial to determine whether agents can faithfully execute human instructions, which is essential for practical use.
To address these issues, we introduce a \sm-CSS, which is better tailored for evaluating multimodal agents.
In \sm-CSS, an agent is expected to strictly take a sequence of actions using our provided toolkit to accomplish a task (Section.~\ref{appendix:css_actions}).
Specifically, it needs to iteratively refine the CSS definition based on the rendering outcomes it receives.
The more constrained action space based on our toolkit, compared to outputting an entire HTML file, along with a more practical goal for current LMMs (i.e., CSS bug-fixing), makes it possible to evaluate a definitive success rate for a given task.
Additionally, \sm-CSS makes minimal assumptions in terms of simplifying the task environment, such as embedding all CSS definitions within a single HTML page or replacing images with placeholders in existing datasets. Instead, the agent directly operates over the entire web frontend project to fix the CSS style.
See a comprehensive checklist in Table~\ref{table:css_related}.

\begin{table}[!h]
\centering
\caption{A fine-grained comparison of \sm-CSS with existing datasets on web frontend development.
\sm-CSS provides both training and test data.
Additionally, its multi-round nature, definitive success rate metric, and multi-file environment make it well-suited as a practical multimodal agent task.
}
\begin{tabular}{lccccc}
\toprule
 & Train & Test  & Multi Round & Definitive Eva. & Multi-File Env. \\
\midrule
WebSight~\cite{websight} & \greencheck  & \redx & \redx & \redx & \redx \\
Design2Code~\cite{design2code} & \redx & \greencheck & \redx & \redx  & \redx \\
\sm-CSS & \greencheck & \greencheck & \greencheck & \greencheck & \greencheck\\
\bottomrule
\end{tabular}
\label{table:css_related}
\end{table}

\subsection{Data Collection}
\label{appendix:css_data}

\begin{figure}[t]
    \centering
    \includegraphics[width=0.9\textwidth]{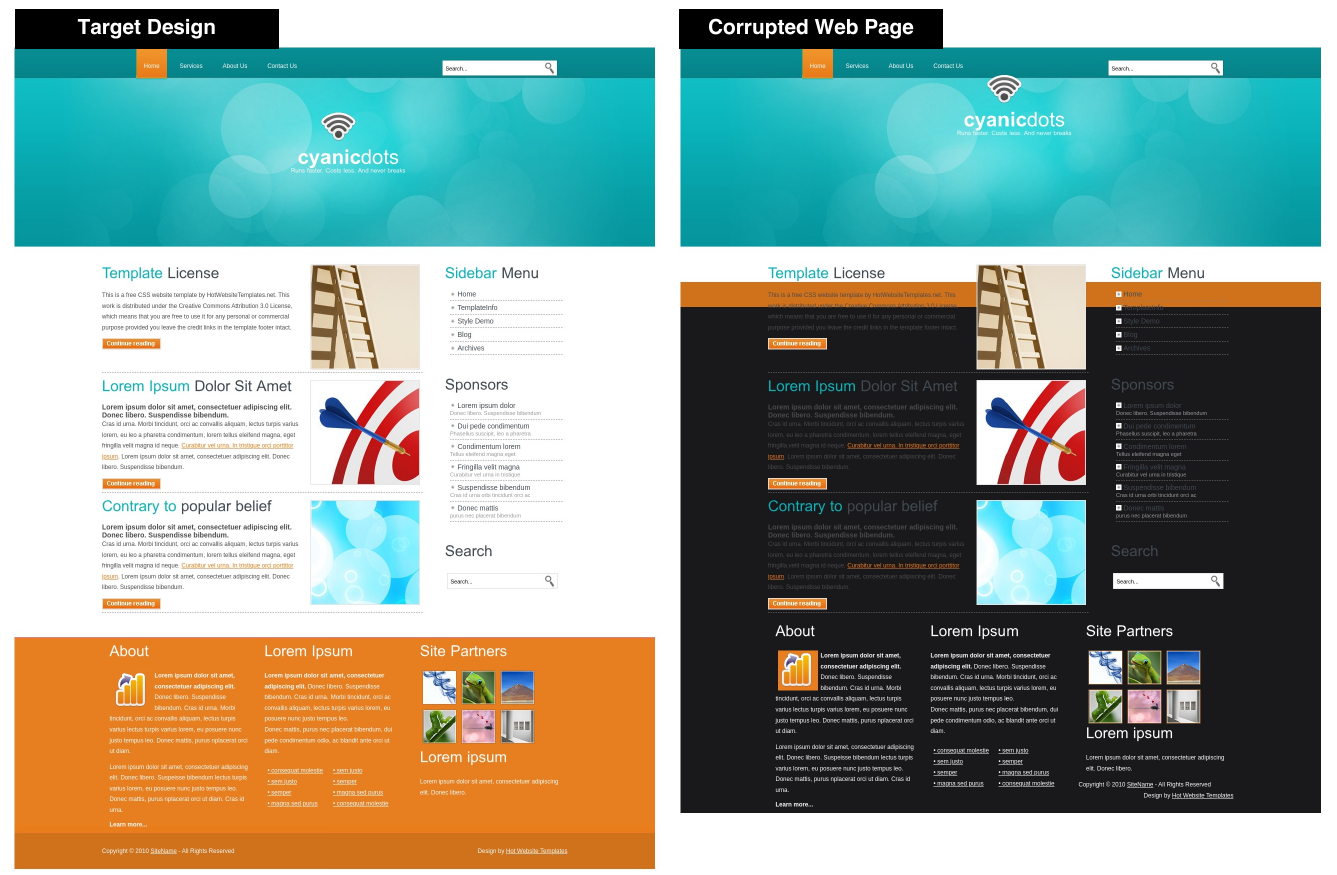}
    \caption{
    This is an example of our annotation task.
    Authors are shown the target design and a corrupted web page side by side to prompt them to describe necessary adjustments in natural language.
    In this example, the instruction is: ``\textit{Correct the background color of the footer and main section, and adjust the positioning of elements, including centering the website logo in the header by moving it downward.}''
    The two screenshots, along with the HTML code and annotated instruction, will collectively serve as the initial task input for the agent.
    }
    \label{fig:css_annotation}
\end{figure}

\vpara{Random CSS Corruption.}
To ensure the task is manageable for LMMs, each task instance involves corrupting a single categorical property of a random CSS rule by either altering its value or removing it entirely.
Note that, even fixing a single corruption is already highly challenging for current LMMs, and a tiny corruption can often lead to a drastic change in visual effect (see Figure~\ref{fig:css_annotation}).
We can increase the task's complexity in the future by involving multiple corruptions once the single-corruption task has been mastered.

\vpara{Human Annotations.}
Existing LMMs struggle to identify the difference between the current rendering and the target design, so we manually annotate each instance with a natural language description of the difference between the two images.
Such natural language descriptions could serve as additional clues for the agent to perceive the visual difference (see a concrete example of annotation in Figure~\ref{fig:css_annotation}).

\vpara{Training Trajectories.}
\label{appendix:css-training}
To collect training trajectories, we primarily sample from the predictions of \texttt{gpt-4o} on our training instances, retaining the successful trajectories for training. 
Given the success rate of \texttt{gpt-4o} is around \num{35}\%, we also sample its trajectories in a more lenient setting where the target CSS rule to edit is provided as input. 
For task instances where \texttt{gpt-4o} succeeds in the lenient setting, we combine its successful trajectory with its failure trajectory in the standard setting to create a more realistic trial-and-error trajectory.

\subsection{Actions}
\label{appendix:css_actions}
In \sm-CSS, the agent is expected to interact with a practical frontend project, potentially with numerous CSS files, to fix its style issues.
Inputting the entire project directly into the agent is impractical and inefficient. 
Instead, the agent has access only to screenshots and the current HTML code.
To facilitate effective navigation and editing within the project, we provide the agent with a toolkit. 
This toolkit allows the agent to locate and edit incorrect CSS definitions seamlessly, without needing to know the specific file containing the CSS rule.

\begin{itemize}[leftmargin=1.5em,itemsep=0pt,parsep=0.2em,topsep=0.0em,partopsep=0.0em]
    \item \textbf{get\_selectors\_by\_html\_element}:
    This function allows the agent to locate a list of CSS selectors, potentially from various files, associated with an HTML element whose style appears to be incorrect.
    \item \textbf{select\_rule}: This function allows the agent to check the definition of a CSS rule by specifying a CSS selector.
    \item \textbf{edit\_rule}: This function enables the agent to update the property value of a CSS rule for a specified CSS selector.
    \item \textbf{revert\_last\_edit}: During the trial and error, the agent can revert an edit it later determines to be incorrect.
\end{itemize}

\subsection{Metrics}
As discussed earlier, a critical feature of \sm-CSS, compared with existing benchmarks, is its definitive success rate evaluation. 
The most straightforward way to determine whether a task is successfully handled is to check whether the SSIM similarity between the target design and the final rendering is \num{1.0}. 
However, we have observed that this can be too strict. 
Typically, an SSIM greater than \num{0.9} indicates minimal differences that are hard for humans to perceive.\footnote{This threshold of \num{0.9} is an empirical choice based on our observations.} 
Therefore, we define a task as successful if the final similarity is greater than \num{0.9}.
Finally, we adopt two metrics on our entire test set.
\begin{itemize}[leftmargin=1.5em,itemsep=0pt,parsep=0.2em,topsep=0.0em,partopsep=0.0em]
    \item \textbf{Success Rate (SR)}:
    This is the primary metric indicating the ratio of tasks in the test set that have been successfully fixed based on our definition.
    \item \textbf{Improve Rate (IR)}: This metric evaluates the ratio of tasks where the final rendering is more similar to the target design than the initial rendering. It serves as a complementary soft metric to the success rate.
\end{itemize}

\subsection{Prompt Example}
\label{appendix:css_prompt}
The system message that describes the detailed task information to the agent is shown as follows:
\lstset{
    backgroundcolor=\color[RGB]{245,245,244},
    breaklines=true,
    basicstyle=\ttfamily\small
}\begin{lstlisting}
You are a CSS agent. You will be given a target screenshot and an html file. Your job is to correct perceive the layout difference between the current rendering and the target screenshot, then accordingly fix the css rules used in the html file to match the target screenshot.
To facilitate the process, you can use the following tools provided by the system:
1. get_selectors_by_html_elements
Sometimes, the exact selector of the rule you want to edit is not clear. This tool takes the html element specification that could be directly passed to soup.find_all as input and returns the matched selectors. For example, get_selectors_by_html_elements("'a', {'data-custom': 'custom-value'}, string='haha', class_='xxx'"). The argument should be the string representation of valid arguments of the find_all method in BeautifulSoup, which means we can directly do eval(f"soup.find_all({argument})"). Please strictly stick to the usage of BeautifulSoup. Make sure the arguments are valid (e.g., the tag name must be wrapped with quotes, attributes should be a dictionary). You can use this tool to first find the selector of the rule of a specific html element whose style you want to fix.
2. select_rule
This takes the css rule's selectorText as input and returns the rule. You can use this tool to view the properties of a rule, which may help you to decide which rule to edit. Usually, it's recommended to first use this tool to view the rule before deciding which rule to edit.
3. edit_rule
This takes the css rule's selectorText, the property name, and the value of the property as input. You can use this tool to change the value of a property of a rule or insert a new property to the rule, if you believe this change would make the rule closer to the target screenshot. Note that, most of the layout issues are related to the categorical properties, such as border, float, display, overflow, position, etc.
4. revert_last_edit
This tool reverts the last single edit you made. You can use this tool to undo the last edit, if you believe it was a mistake. This action takes no arguments.

Make sure the selectorText is valid based on the html file, i.e., it's from the class or id of the html elements. In addition, please focus on the major layout issue! Ignore the font size, font family, and color of the text, even if you believe they are not perfect.

You can only take ONE action at a time!! For each step, you may first state your thought, then take an action following the format of Thought: ...
 Action: ... (do not add any linebreak after the colon).
For example, you may output
"Thought: I think I should adjust the alignment property of the rule, because the target screenshot shows the text should be centered.
 Action: edit_rule('.templatemo_menu li', 'text-align', 'center')".

After editing a rule or inserting a rule, you will see the updated screenshot of the html file. You should decide your next action (e.g., to revert the last edit or keep adjusting the css) based on the updated screenshot. If you think you have already fixed the css style, please say exactly "I have fixed the css style".

Please strictly follow the format specified above, and please don't repeat the same action in multiple rounds. Also note that, you don't need to worry about how these tools are executed, your job is just to correctly predict the tool invocation.

\end{lstlisting}

Here is a concrete example of the task input shown in Fig.~\ref{fig:css_annotation}, where variables are enclosed within ``\{\{\}\}'':
\lstset{
    backgroundcolor=\color[RGB]{245,245,244},
    breaklines=true,
    basicstyle=\ttfamily\small
}\begin{lstlisting}
Here is a screenshot of the target design:
{{Image 1}}
Here is the screenshot of the current web page:
{{Image 2}}
Here is the HTML code of the current web page:
{{HTML file}}

Correct the background color of the footer and main section, and adjust the positioning of elements, including centering the website logo in the header by moving it downward.
\end{lstlisting}

\end{document}